\documentclass[pdflatex,sn-mathphys-ay]{sn-jnl}

\usepackage{graphicx}%
\usepackage{multirow}%
\usepackage{multirow}
\usepackage{amsmath,amssymb,amsfonts}%
\usepackage{amsthm}%
\usepackage{mathrsfs}%
\usepackage[title]{appendix}%
\usepackage{xcolor}%
\usepackage{textcomp}%
\usepackage{manyfoot}%
\usepackage{booktabs}%
\usepackage{algorithm}%
\usepackage{algorithmicx}%
\usepackage{algpseudocode}%
\usepackage{listings}%
\usepackage[table]{xcolor}
\usepackage{adjustbox}
\theoremstyle{thmstyleone}%

\theoremstyle{thmstyletwo}%

\theoremstyle{thmstylethree}%
\begin{document}

\title{Social media polarization during conflict: Insights from an ideological stance dataset on Reddit comments}

\author[1]{\fnm{Hasin Jawad} \sur{Ali}}\email{hasinjawad@iut-dhaka.edu}

\author*[2]{\fnm{Ajwad} \sur{Abrar}}\email{ajwadabrar@iut-dhaka.edu}

\author[3]{\fnm{S.M. Hozaifa} \sur{Hossain}}\email{20hozaifa02@gmail.com}

\author[4]{\fnm{M. Firoz} \sur{Mridha}}\email{firoz.mridha@aiub.edu}

\affil[1]{\orgdiv{Department of Business and Technology Management},
\orgname{Islamic University of Technology},
\orgaddress{\city{Gazipur}, \state{Dhaka}, \country{Bangladesh}}}

\affil[2]{\orgdiv{Department of Computer Science and Engineering},
\orgname{Islamic University of Technology},
\orgaddress{\city{Gazipur}, \state{Dhaka}, \country{Bangladesh}}}

\affil[3]{\orgdiv{Department of Electrical and Electronic Engineering},
\orgname{University of Dhaka},
\orgaddress{\street{Nilkhet Road}, \city{Dhaka}, \country{Bangladesh}}}

\affil[4]{\orgdiv{Department of Computer Science},
\orgname{American International University–Bangladesh},
\orgaddress{\street{Kuratoli}, \city{Dhaka}, \country{Bangladesh}}}

\abstract{In politically sensitive scenarios such as wars, social media often becomes a platform for polarized discourse and expressions of strong ideological stances. While prior studies have explored ideological stance detection in general contexts, limited attention has been given to conflict-specific settings. This study addresses this gap by analyzing 9,969 Reddit comments related to the Israel–Palestine conflict collected between October 2023 and August 2024. The comments were manually annotated into three stance classes—Pro-Israel, Pro-Palestine, and Neutral—with high inter-annotator agreement (Fleiss' Kappa = 0.93). We conduct a comprehensive comparative evaluation of neural networks, pre-trained language models, and large language models (LLMs), along with multiple prompt engineering strategies and an LLM majority voting ensemble. Model performance is assessed using accuracy, precision, recall, and F1-score. Among the tested approaches, the Scoring and Reflective Re-read prompting strategy applied to Mixtral 8x7B achieves the best overall performance across evaluation metrics. Beyond model performance, the findings provide empirical insights into how ideological polarization manifests in online discourse surrounding the Israel–Palestine conflict, offering a computational perspective on the distribution of Pro-Israel, Pro-Palestine, and Neutral narratives. The dataset used in this study is publicly available to support further research on ideological stance detection in politically sensitive contexts.}

\keywords{Ideological Stance Detection, Machine Learning, Large Language Models, Israel-Palestine Conflict, Social Media Analysis}

\maketitle
\section{Introduction}
Ideological stances on the part of one or both parties can increase the likelihood of conflicts between two nations. If the stances are sufficiently strong, they even render mutually acceptable transfers insufficient to prevent war \citep{jackson2007political}. And when a war has already broken out, these stances come to surface extensively on social media. \cite{kubin2021role} states that social media reinforce these existing positions through the fragmentation of the news media and facilitate the spread of misinformation. In the context of recent Israel-Palestine conflict, social media is a hub for the exchange of opposing political ideologies. And due to the historical nature of this conflict, this manifestation is often highly polarized from both sides.

Numerous studies have been conducted in order to develop methodologies for identifying polarizing stances. Those studies have employed numerous machine learning algorithms, ranging from classical supervised algorithms to neural networks to pre-trained language models. Large language models (LLMs) have been used in sentiment detection of political texts \citep{kuila2024deciphering}. A study \citep{kwon2024sentiment} has also compared LLMs, neural networks, pre-trained language models and classical supervised algorithms in the context of ideological sentiment detection. Another study \citep{ansari2020analysis} compares LSTM and classical supervised algorithms in ideological stance detection. However, no studies have yet been conducted that use LLMs in ideological stance detection or compare LLMs to neural networks or pre-trained language models in this context. Moreover, no studies have also applied prompt engineering techniques in this context. 

To address this research gap, we conduct a comprehensive comparative analysis of multiple approaches for ideological stance detection, including open-access large language models (LLMs) with various prompt engineering techniques, neural network architectures, and pre-trained language models. We further identify the best-performing method based on multiple evaluation metrics. Specifically, we evaluate seven open-access LLMs (Mixtral 8x7B, Mistral 7B, Gemma 7B, Falcon 7B, Gemma 3-4B, Falcon 3-3B, and Qwen 3-4B), four neural network architectures (RNN, LSTM, GRU, and BiLSTM), and seven pre-trained language models (BERT Cased, BERT Uncased, XLM-RoBERTa, DistilBERT Cased, DistilBERT Uncased, ELECTRA-Small, and ELECTRA-Base). For LLM-based classification, we investigate several prompt engineering strategies, including Zero-shot prompting, One-shot prompting, Re2 prompting, Meta-prompting, Scoring and Reflective Re-read prompting, and Context Extraction prompting, and we also examine an LLM ensemble approach based on majority voting. We utilized a manually annotated dataset of Reddit comments related to the Israel-Palestine conflict, spanning from October 2023 to August 2024. The models were evaluated based on accuracy, F1-score, recall, and precision on the test dataset. The final methodology was selected by systematically weighing these performance metrics. This comprehensive evaluation not only highlights the capabilities of various models in detecting ideological stance but also identifies the most suitable approach for this task. The contributions of this study are as follows:

\begin{itemize}
    \item A meticulously manually annotated and publicly available dataset containing the ideological stances of \textbf{9,969 Reddit comments} in the context of the Israel-Palestine conflict, with high inter-annotator agreement (\textbf{Fleiss' Kappa = 0.93}).\footnote{The data is publicly available at \url{https://github.com/jami78/Conflict-Bias-Eval}}
    
    \item A comprehensive comparative analysis of \textbf{18 models}, including 7 pre-trained language models, 4 neural network architectures, and 7 large language model configurations, for ideological stance detection in politically sensitive and highly polarized scenarios.
    
    \item The establishment of benchmark models and the development of \textbf{novel prompt engineering strategies} along with an \textbf{LLM majority voting ensemble}. Among the evaluated approaches, the proposed Scoring and Reflective Re-read prompting strategy achieves the best overall performance across evaluation metrics.
\end{itemize}

\label{sec1}

The remainder of the paper is organized as follows:  Section \ref{rw} provides a comprehensive review of the existing literature; Section \ref{data} outlines the details of the dataset; Section \ref{method} demonstrates the proposed methodology including the models that were used; Section \ref{es} details the hardware setup and evaluation metrics used in our experiments; Section \ref{pe} presents the performance scores of the models; Section \ref{bi} analyzes the broader implications of this research and finally, Section \ref{conclusion} provides a conclusion with a summary of the findings and a few potential future directions.
\section{Related Work} \label{rw}
Ideological stance detection in our context refers to the identification of the underlying ideological viewpoint that a text might have conveyed in response to a particular political scenario. Traditionally, this domain used to leverage classical machine learning techniques until the emergence of neural networks and pre-trained language models significantly changed the landscape \citep{AlSarraj2018},\citep{ansari2020analysis}. More recently, large language models (LLMs) have advanced this field with the help of several prompt engineering strategies due to their ability to capture textual nuances that the previous models failed to do \citep{kuila2024deciphering}, \citep{mouthami2025}. This section reviews the progression of methodologies in ideological stance detection, organized into three categories: neural networks, pre-trained language models, and state-of-the-art LLMs. The previous studies of stance detection in the context of the Israel-Palestine conflict are also discussed. 

\subsection{\textbf{Neural Networks}}
Neural networks emerged as an improvement over the classical supervised model in the domain of ideological stance detection as it could capture patterns better. For example, \citep{ansari2020analysis} demonstrated that LSTM models outperform classical supervised models in ideological stance detection due to their ability to capture the temporal dependencies and context of tweets as observed in its performance on the Twitter dataset for 2019 Indian General Elections. According to \citep{khatua2020predicting}, Bi-LSTM is even better than LSTM, RNN, and SVM with a motion-dependent debate model for this particular task. When an RNN model is enhanced with word embeddings (W2V), it outperforms traditional Logistic Regression that uses Bag of Words (BoW) and word embeddings (W2V) \citep{iyyer2014political}. \citep{hamborg2021news} proposes a BiGRU model with multiple embeddings that outperformed previous state-of-the-art models for target-dependent sentiment classification in news articles. The model's success can be attributed to its ability to handle longer texts and multi-target sentences, offering a better solution for real-world sentiment analysis tasks in news media.

\subsection{Pre-trained Language Models}
Pre-trained language models further improved performance over the neural networks by offering better handling of context. They have been used extensively over the years in ideological stance detection. According to \citep{ozturk2024ideology}, BERT demonstrates superior performance compared to classical supervised models, LSTM, and BiLSTM in this domain. However, \citep{abercrombie2020parlvote} reported that a supervised SVM classifier with a motion-dependent debate model outperformed the BERT+MLP model in detecting ideological sentiment in a large UK parliamentary debate corpus known as ParlVote. Furthermore, ELECTRA, another pre-trained language model, has been shown to outperform BERT in the detection of ideological stance \citep{ozturk2024ideology}. BERT, which utilizes transfer learning and logistic regression, has outperformed rule-based VADER and TextBlob in the detection of sentiment related to Sustainable Development Goals (SDGs) \citep{rosenberg2023sentiment}. BERT-based models have also been shown to be more socially conservative compared to GPT models, effecting their classification performance downstream. As a result, these models might exhibit certain biases, particularly in hate speech detection, unless they are pre-trained on left-leaning or right-leaning corpora \citep{feng2023from}. In a separate study, \citep{abrar2025religious} concluded that language models such as BERT, RoBERTa, and DistilBERT continue to exhibit significant religious biases. RoBERTa-Base has also performed better than LSTM, VADER, XLM-R, and SVM in monolingual ideological sentiment classification \citep{antypas2023negativity}. However, according to \citep{ferracane2023integrated}, a RoBERTa-based model specifically fine-tuned for classification in the political domain, named POLITICS, outperformed RoBERTa-Base in terms of detecting ideological stance by leveraging discourse-based, structure-aware representations. Another multi-task learning model named CLoSE, introduced in \citep{kim2022close}, was shown to outperform both RoBERTa and BERT due to its capability of embedding sentence framing language and predicting ideological stance simultaneously. The study \citep{bhatia2018topic} showed that the TSM (Topic-Sentiment Matrix) model performed comparably to state-of-the-art distributional text representation models such as GloVe-d2v, achieving an accuracy of 65.04\% compared to 64.30\% with GloVe-d2v. DistilBERT has shown better performance than ELMo in the classification of protest news and sentiment analysis in cross-context settings \citep{buyukoz2020analyzing}.

\subsection{Large Language Models (LLMs)}
Large language models (LLMs) are the latest innovation in stance detection which are capable of understanding contexts and textual nuances better than any previous model. For example- LLMs showed better classification performance than BERT, CNN, traditional supervised algorithms and neural networks in ideological sentiment classification \citep{kwon2024sentiment}. Similarly, \citep{kuila2024deciphering} evaluated the performance of several open-access large language models (LLMs) in detecting ideological sentiments which identified Falcon 40B as the best-performing model, followed by Mistral 7B and Falcon 7B. However, LLMs are not without challenges. LLMs have been shown to have a left-leaning tendency, which might affect their classification accuracy \citep{hernandes2024llms, pit2023whose}. Additionally, their high computational requirements might restrict their efficiency. 

\subsection{\textbf{Stance Detection in the Israel-Palestine Conflict}}
The Israel-Palestine war has instigated a few ideological stance detection studies, each with its own strengths and limitations. \citep{AlSarraj2018} investigated media stance in the coverage of the 2014 Israeli war on Gaza by Western news outlets. The study employed text mining techniques and supervised machine learning algorithms, such as SVM, to classify stances into binary categories of Pro-Palestine and Pro-Israel. The models showed high performance, achieving an accuracy and recall of approximately 90\% in classifying these ideological stances. However, this study did not consider neutrality as a stance category, which limits its capacity to capture balanced stances in politically sensitive content. On the other hand, \citep{imtiaz2022takingsidespublicopinion} incorporated neutrality as a stance category, broadening the range of perspectives considered. The study also achieved incredible model performance using XLNet. Nonetheless, it utilized a Twitter dataset, which may have limited the text's ability to capture nuanced expressions due to the 280-character limit. Additionally, the study did not include state-of-the-art large language models (LLMs) for its analysis.

Building on this foundation, our work broadens the scope by integrating neural networks, pre-trained language models, and open source large language models (LLMs) with prompt engineering to analyze Reddit comments related to the Israel-Palestine conflict. Unlike traditional machine learning techniques, these advanced models enable a deeper understanding of contextual nuances, leading to the dynamic classification of ideological stance into three categories: Pro-Palestine, Pro-Israel, and Neutral. Although \citep{kuila2024deciphering} applied several existing prompting techniques for political sentiment classification, they did not introduce any novel methods for superior stance detection performance. We address this by evaluating multiple novel prompting strategies and proposing a new technique that improves stance detection on a manually annotated Reddit dataset of the Israel-Palestine conflict, which captures longer and more nuanced discourse than typical Twitter \citep{imtiaz2022takingsidespublicopinion} and non-conflict datasets.

\section{Dataset} \label{data}
The dataset used in this study is a filtered version of the 'Daily Public Opinion on Israel-Palestine War' dataset \citep{asaniczka2024daily}, obtained from Kaggle. It contains comments from Reddit posts related to the ongoing Israel-Palestine war collected over a year from September 2023 to September 2024. This dataset has an advantage over previous datasets used for ideological stance detection because the context is set during a politically charged situation.
\subsection {\textbf{Dataset Comparison}}
Table \ref{tab:dataset_comparison} illustrates the difference between the previous datasets and ours. Our dataset has an advantage over them due to the context being set during a politically charged situation. It is also manually annotated to a greater degree than the previous ones.

\begin{table*}[tbp]
\centering
\caption{Comparison of related datasets}
\label{tab:dataset_comparison}

\resizebox{\textwidth}{!}{%
\begin{tabular}{|p{2.6cm}|p{2.3cm}|p{2.6cm}|p{1.2cm}|p{2.0cm}|p{2.5cm}|p{3.8cm}|}
\hline
\textbf{Dataset} & \textbf{Platform} & \textbf{Conflict/Topic} & \textbf{Year} &
\textbf{Size} & \textbf{Stance Classes} & \textbf{Limitations} \\ \hline

Imtiaz et al. (2022) \citep{imtiaz2022takingsidespublicopinion}
& Twitter & Israel-Palestine Conflict & 2021
& 493,877 tweets
& Pro-Palestine, Pro-Israel, Neutral
& Short-form tweets; limited manual annotation \\ \hline

Al-Sarraj et al. (2018) \citep{AlSarraj2018}
& News websites & Israel-Palestine Conflict & 2014
& 799 articles
& Pro-Palestine, Pro-Israel
& No neutral class; limited time span \\ \hline

Ansari et al. (2020) \citep{ansari2020analysis}
& Twitter & Indian General Elections & 2019
& 3,896 tweets
& Multi-party labels
& Not conflict-focused \\ \hline

Hernandez et al. (2022) \citep{hernandes2024llms}
& News articles & U.S. political bias & --
& 5,877 articles
& Multi-level ideology
& Outlet-based labels \\ \hline

Bhatia et al. (2018) \citep{bhatia2018topic}
& Parliamentary debates & U.S. ideology & 2005
& 1,586 docs
& Republican, Democrat
& Two-party only \\ \hline

\textbf{Ours}
& \textbf{Reddit}
& \textbf{Israel-Palestine Conflict}
& \textbf{2023--2024}
& \textbf{9,969 comments}
& \textbf{Pro-Palestine, Pro-Israel, Neutral}
& \textbf{Conflict-specific} \\ \hline

\end{tabular}%
}
\end{table*}

\subsection{\textbf{Data Collection and Filtering}}
The original dataset comprises 1,825,545 comments collected from various subreddits. In this study, we used keyword filtering, a widely used NLP technique for summarization and retrieval, in order to identify contextually relevant comments as summarized by \citep{chen2024isamasredpublicdatasettracking}. This filtering process resulted in a refined subset of 9,969 comments focused on terms such as ``Hamas terrorist”, “Zionist terrorists”, “Stand With Israel”, ``Free Palestine", ``October 7", ``From the river to the sea", ``Jewish state", ``Hostage", and ``Israel occupation”. This sample size strikes a balance between representativeness (ensuring that the comments capture complete and diverse expressions) and feasibility (enabling high-quality manual annotation and systematic LLM evaluation). Subsequently, duplicate comments were removed to ensure that the dataset consisted of unique and unambiguous entries.

\subsection{\textbf{Data Annotation}}
After filtering the comments, they were manually annotated for sentiment by three independent annotators. Each comment was assigned one of the three ideological stance labels: Pro-Israel, Pro-Palestine, or Neutral. The final label for a comment was determined based on majority agreement among the annotators; if two or more annotators assigned the same label, it was accepted as the final label. Notably, no instances occurred in which all three annotators assigned different labels to the same comment.

The process utilized an initial filtering process using specific keywords before annotation:
\begin{itemize}
    \item \textbf{Pro-Israel:} Comments containing keywords that expressed support for Israel and opposition to Palestine or Hamas, such as \textit{“Hamas terrorist”, “Stand With Israel”, “Hamas terror”, “Support Israel”, “Israel Strong”, “Israel Forever”, “Israel Under Attack”}.
    \item \textbf{Pro-Palestine:} Comments containing keywords that expressed support for Palestine and opposition to Israel, such as \textit{“Zionist terrorists”, “Zionist terror”, "Free Palestine", “Boycott Israel”, “Palestine will be free”}.
    \item \textbf{Neutral:} Comments that did not convey a strong sentiment towards either side were categorized as Neutral.
\end{itemize}

This systematic annotation process ensured that the dataset was accurately labeled for stance detection. This annotation process resulted in a total of 2,431 comments labeled Neutral, 4,947 comments labeled Pro-Israel, and 2,591 comments labeled Pro-Palestine. This balanced distribution of labels reflects a diverse range of stances and sentiments, ensuring a comprehensive dataset for analyzing ideological stance.

To measure inter-annotator agreement, we utilized Fleiss' Kappa \citep{fleiss1971measuring}. The calculated score for the three annotators is 0.93, indicating a high level of agreement.

\subsection{\textbf{Dataset Statistics}}
\begin{figure}
    \centering
    \includegraphics[width=1\linewidth]{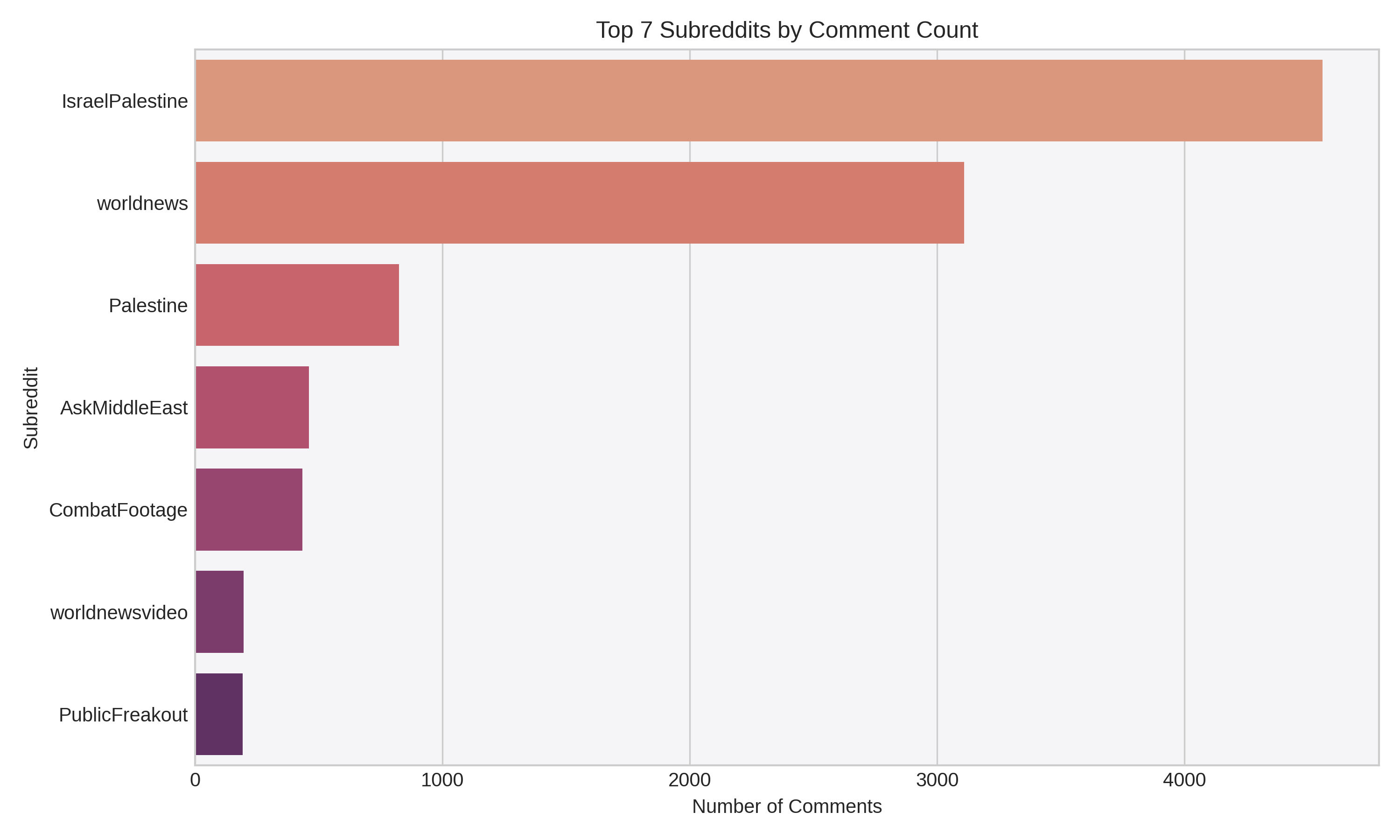}
    \caption{Top 7 Subreddits}
    \label{fig:subreddits}
\end{figure}

Table \ref{tab:dataset_overview} shows dataset statistics. There are a total of 5589 users with 1.78 average comment each. The entire dataset contains a total of 28,116 unique words and 12 unique subreddits. 
Among the subreddits, r/IsraelPalestine, r/worldnews, and r/Palestine are the most frequent as shown in figure \ref{fig:subreddits}. 

\begin{table}
\centering
\caption{Dataset Overview}
\label{tab:dataset_overview}
\begin{tabular}{@{}lc@{}}
\toprule
\textbf{Metric} & \textbf{Value} \\
\midrule
Total comments & 9,969 \\
Unique users & 5,589 \\
Comments per user (mean) & 1.78 \\
Total words (tokens) & 1,149,661 \\
Unique words (vocabulary size) & 28,116 \\
Unique subreddits & 12 \\
\bottomrule
\end{tabular}
\end{table}

\begin{figure}
    \centering
    \includegraphics[width=1.0\linewidth]{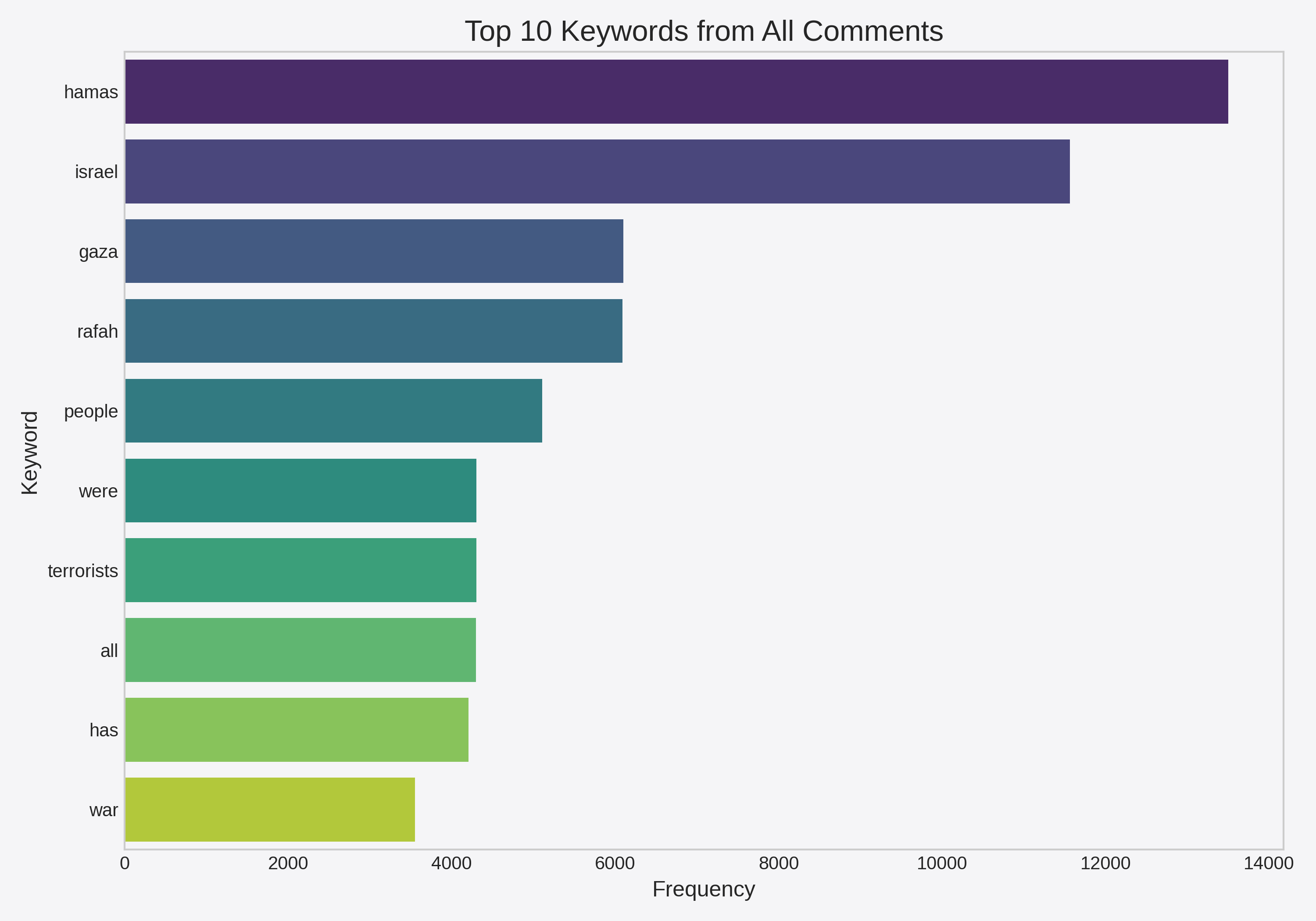}
    \caption{Top 10 Keywords}
    \label{fig:keywords}
\end{figure}

\begin{figure}
    \centering
    \includegraphics[width=1\linewidth]{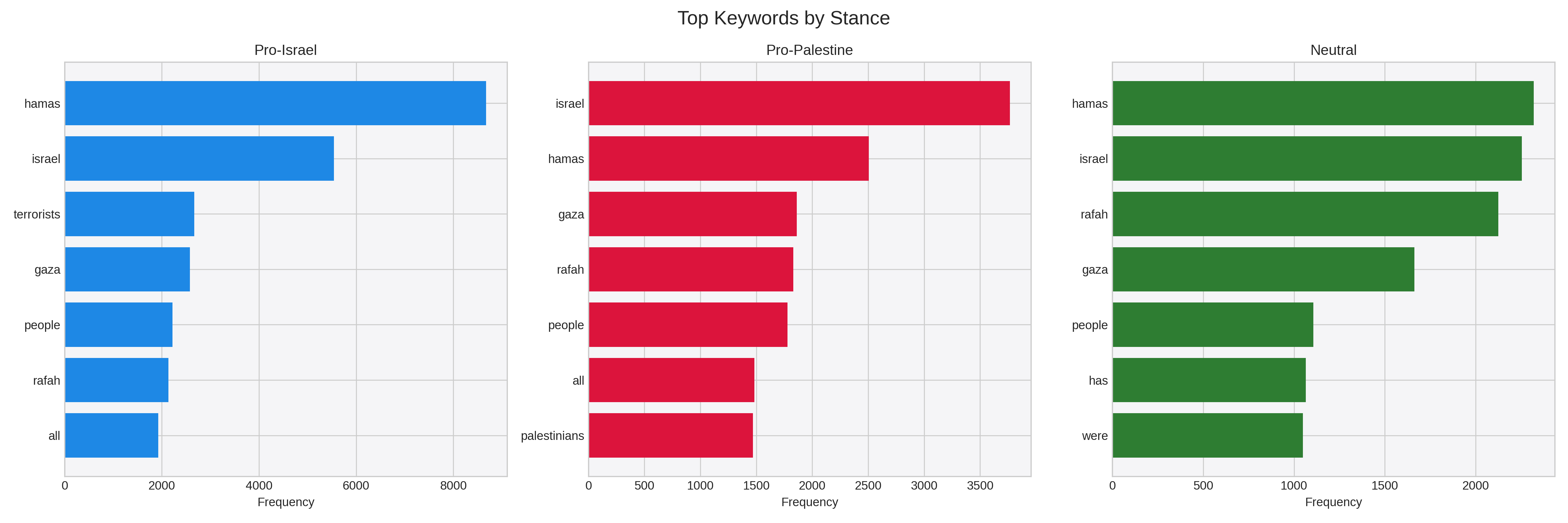}
    \caption{Keywords Per Stance}
    \label{fig:keyword_stance}
\end{figure}

According to Figure~\ref{fig:keywords}, \textit{“hamas”}, and \textit{“israel”}, emerge as the most frequently used keywords, surpassing others by a significant margin. Interestingly, these two words are also the most frequently used keywords in the texts corresponding to all three stances (Figure ~\ref{fig:keyword_stance}). However, the word \textit{“terrorist”} appears significantly more in Pro-Israel stances than the other two. The word \textit{“gaza”} also appears regularly in all three stances. 

\begin{figure}[ht]
\centering
\includegraphics[width=1.0\textwidth]{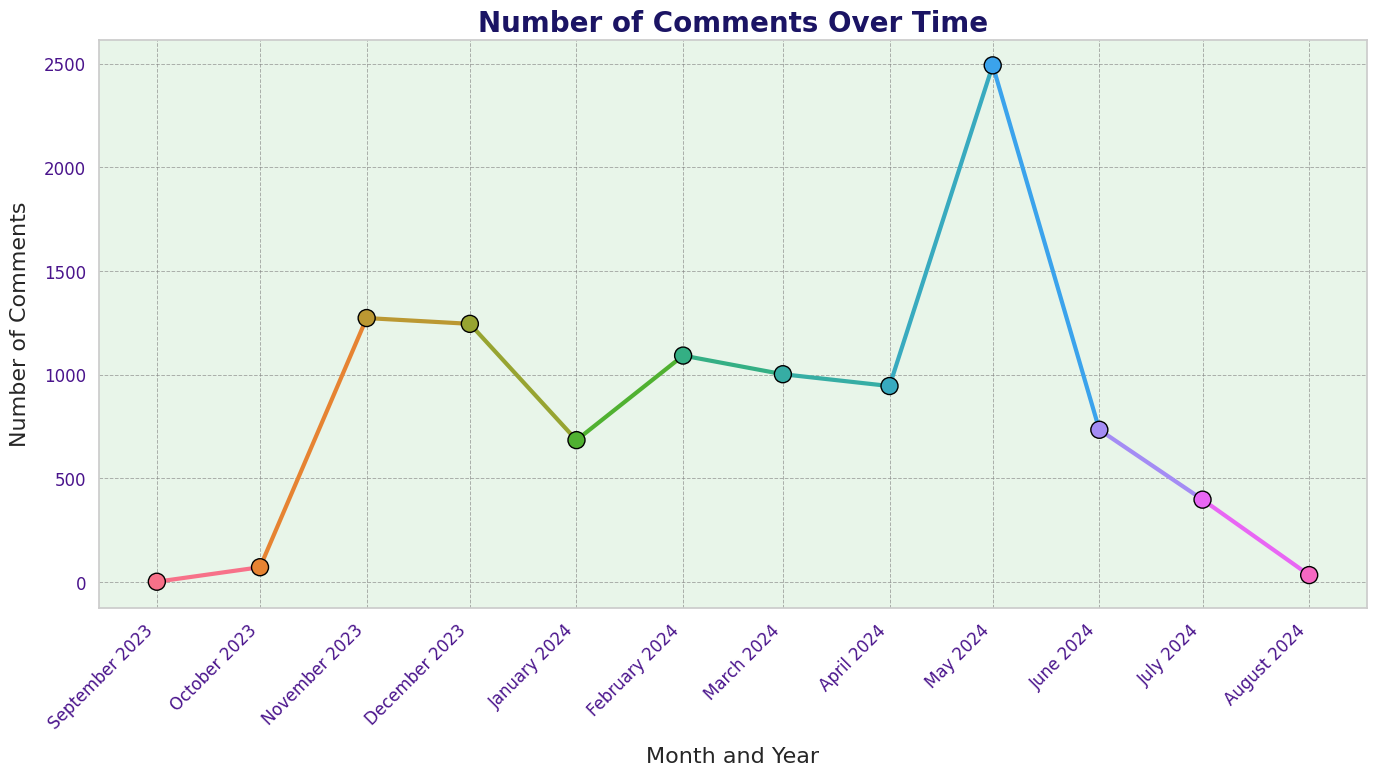}
\caption{Time Distribution of Comments}
\label{fig:time-distribution}
\end{figure}

\begin{figure}
    \centering
    \includegraphics[width=1\linewidth]{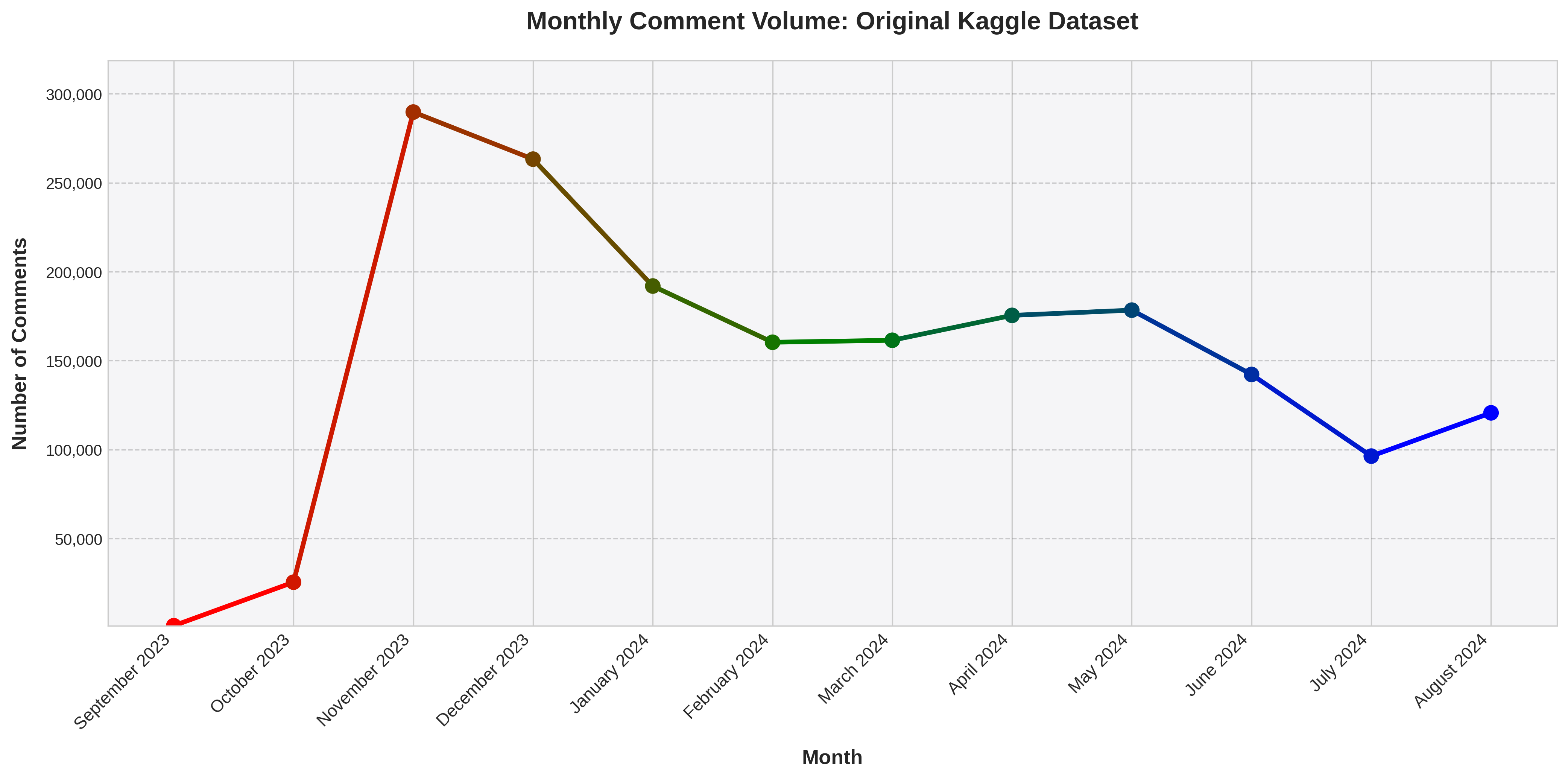}
    \caption{Monthly Comment Volume of the Original Dataset}
    \label{fig:original}
\end{figure}

\begin{figure}
    \centering
    \includegraphics[width=1\linewidth]{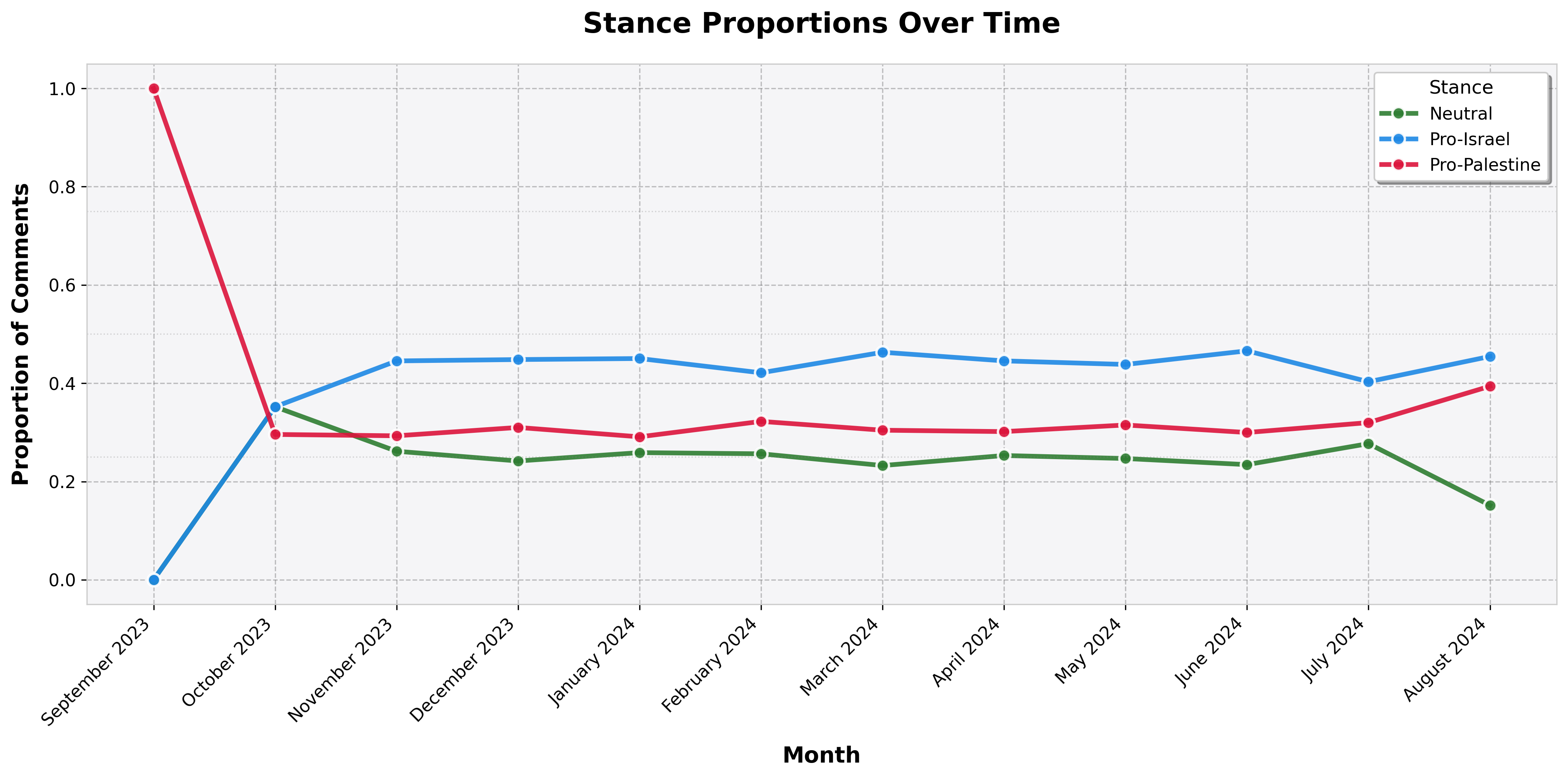}
    \caption{Stance Proportions Over Time}
    \label{fig:stance_proportion}
\end{figure}

The temporal distribution of comments (Figure~\ref{fig:time-distribution}) reveals that the number of comments is generally centered around 1,000 per month, with the highest volume observed in May 2024 and the lowest during the first and last two months of the dataset's time span. The original, unfiltered dataset does not show the exact same trend since November 2023 contains the highest number of comments according to figure \ref{fig:original}. In terms of stance proportion per month, Pro-Palestine had highest number in September 2023 which dipped below the Pro-Israel level in October 2023. After that, the proportion remained consistent (Figure \ref{fig:stance_proportion}).


\subsection{\textbf{Data Processing}} 
To prepare the dataset for model training and evaluation, a series of preprocessing steps were performed. These steps ensured that the data were clean, consistent, and in a format suitable for the application of machine learning and deep learning models. The following key techniques were employed during the data processing phase:

\begin{itemize}
    \item \textbf{Label Encoding:} Sentiment labels were converted into a numerical format for compatibility with neural networks: Neutral = 0, Pro-Palestine = 1, and Pro-Israel = 2.
    
    \item \textbf{Text Vectorization:} In order to make the data suitable for neural networks, multiple vectorization techniques were employed to transform text into numerical representations. These included Bag of Words (BoW), Term Frequency-Inverse Document Frequency (TF-IDF), FastText, and Word2Vec. The most effective technique for each model type was experimentally determined to optimize performance.
    
    \item \textbf{Truncation:} To manage computational efficiency and memory usage, texts were truncated to a maximum length of 700 characters, aligning with recommendations for handling lengthy sequences in transformer-based models \citep{devlin2019bert}. Truncated texts were used in all models. 
    
\end{itemize}

\subsection{\textbf{Dataset Split}}

The dataset was randomly divided into three subsets to facilitate model training, evaluation, and validation. Specifically, 6,978 samples (70\%) were allocated for training, 1,496 samples (15\%) for testing, and 1,495 samples (15\%) for validation. This 70:15:15 split ensured a balanced distribution of sentiment labels across all subsets, maintaining consistency for reliable performance assessment.

\section{Methodology} \label{method}
Our study utilized various Neural Networks, pre-trained language models and open source large language models (LLMs) to identify the ideological stances (Pro-Palestine, Pro-Israel, or Neutral) of Reddit comments. The following subsections provide a detailed description of the models used.
\subsection{\textbf{Neural Network Architectures}}

Deep learning models, including the RNN, LSTM \citep{Sherstinsky_2020}, Bi-LSTM \citep{cui2019deepbidirectionalunidirectionallstm}, and GRU \citep{dey2017gru}, were trained and evaluated for the task. Among the various vectorization techniques explored, Bag of Words (BoW) has emerged as the most effective for these architectures. To enhance model performance, hyperparameter tuning was conducted, focusing on optimizing the configurations of the neural network architectures. Table \ref{tab:hyperparameter_ranges} lists the tested hyperparameters and their respective ranges.

\begin{table*}[tbp]
\centering
\caption{Hyperparameter Ranges for Neural Network Architectures}
\label{tab:hyperparameter_ranges}

\resizebox{\textwidth}{!}{%
\begin{tabular}{@{}lccccc@{}}
\toprule
 & \multicolumn{5}{c}{\textbf{Hyperparameters}} \\ 
\cmidrule(lr){2-6}
\textbf{Model} & \textbf{Dropout Rate} & \textbf{Number of Layers} & \textbf{Units per Layer} & \textbf{Batch Size} & \textbf{Learning Rate} \\ 
\midrule
Bi-LSTM & 0.2, 0.3, 0.5 & 1, 2, 3 & 64, 128, 256 & 16, 32, 64 & 0.001, 0.01, 0.1 \\
LSTM    & 0.2, 0.3, 0.4 & 1, 2    & 64, 128      & 16, 32     & 0.001, 0.01 \\
Bi-GRU  & 0.2, 0.3, 0.5 & 1, 2, 3 & 64, 128, 256 & 16, 32, 64 & 0.001, 0.01, 0.1 \\
RNN/GRU & 0.2, 0.3, 0.5 & 1, 2, 3 & 64, 128, 256 & 16, 32, 64 & 0.001, 0.01, 0.1 \\
\bottomrule
\end{tabular}%
}

\end{table*}

\subsection{\textbf{Pre-trained Language Models}}
Pre-trained language models were fine-tuned on the dataset to create task-specific architectures that act as the mediator between traditional neural networks and large language models (LLMs). These models effectively capture text patterns and understand the context and relationships within a text. In addition, they are computationally more efficient than LLMs and can be easily adapted for specific tasks through fine-tuning. In our study, seven fine-tuned models were used to evaluate the task performance.
\begin{itemize}
    \item \textbf{BERT (Cased and Uncased):} BERT is a transformer-based model which is pre-trained using a masked language model objective in order to capture bidirectional context in text. The cased version preserves case sensitivity, whereas the uncased version ignores it \citep{devlin2019bert}.
    \item \textbf{XLM-RoBERTa:} XLM-RoBERTa extends RoBERTa to cross-lingual tasks by leveraging massive multilingual datasets that allow it to achieve enhanced performance on multilingual tasks ranging across 100 languages \citep{conneau2020xlmr}.
    \item \textbf{DistilBERT (Cased and Uncased):} This is the distilled version of BERT which offers faster processing speeds and reduced model size while retaining 97 \% of BERT's performance \citep{sanh2019distilbert}.
    \item \textbf{ELECTRA (Small and Base):} This model employs a unique token-replacement detection mechanism for efficient pre-training which enables the model to train faster and learn from smaller datasets \citep{clark2020electra}.
\end{itemize}

\subsection{Large Language Models}
In politically sensitive datasets such as ours, nuanced language patterns and implicit stances often go unnoticed by traditional models. However, large language models (LLMs) offer a promising solution, as they are pre-trained on extensive and diverse datasets, enabling them to generalize effectively to new contexts with minimal task-specific fine-tuning. Their ability to handle zero-shot, one-shot, and few-shot scenarios makes them particularly advantageous in low-resource settings where labeled data are limited. Additionally, LLMs are well-suited for processing longer and more complex narratives, which is especially relevant for our dataset comprising detailed Reddit comments with subtle political leanings. In this study, we conducted experiments using seven LLMs, all accessed via the Huggingface platform.\footnote{Huggingface platform: \url{https://huggingface.co}}

\begin{itemize}
    \item \textbf{Mistral:} Mistral is a dense and efficient transformer-based language model with 7 billion parameters. It is trained on a mixture of high-quality datasets using techniques such as group-query attention and multi-query key-value caching for efficiency. It demonstrates advanced reasoning and comprehension capabilities, often outperforming Llama 2 13B on standard benchmarks \citep{jiang2023mistral7b}.
    \item \textbf{Mixtral:} Mixtral 8x7B is a Sparse Mixture of Experts (SMoE) model based on Mistral 7B which uses 8 experts per layer while activating only 13B parameters at a time \citep{jiang2024mixtralexperts}.
    \item \textbf{Gemma:} Gemma is a family of lightweight open models based on the Gemini framework that incorporates multi-modal and multi-task training \citep{gemmateam2024gemmaopenmodelsbased}. It offers multiple versions including Gemma 2B, Gemma 7B, Gemma 2-9B, Gemma 2-27B, Gemma 3-4B etc. We used Gemma 7B and Gemma 3-4B in our study which contain 8.54 and 4 billion parameters respectively. 
    \item \textbf{Falcon:} Falcon is a series of causal decoder-only models that includes many versions including Falcon 7B, Falcon 40B, Falcon 180B, Falcon 2-11B, Falcon 3-3B etc., trained on over 3.5 trillion tokens of high-quality web data \citep{Almazrouei2023TheFS}. We utilized Falcon 7B and Falcon 3-3B in our study.
    \item \textbf{Qwen:} The Qwen series, developed by Alibaba Group, comprises Transformer-based large language models trained on diverse multilingual corpora encompassing web texts, books, and code \citep{bai2023qwentechnicalreport}. There are many types of Qwen models including QWEN-CHAT, CODE-QWEN, CODE-QWEN-CHAT,  and MATH-QWEN-CHAT. They also offer many parameter choices including 1,7B, 4B, 122B etc. We utilized Qwen 3-4B in our study. 
\end{itemize}

\subsubsection*{\textbf{Prompting Strategies}}
We applied zero-shot, one-shot, three-shot, and five-shot prompting strategies \citep{brown2020languagemodelsfewshotlearners} to the truncated text column of the dataset to identify the underlying ideological stance in each entry. These strategies evaluated the models' abilities to handle varying levels of contextual information and infer classifications for the text.
\begin{itemize}

    \item \textbf{Zero-Shot Prompting:} In the zero-shot setting, the models were not provided with contextual information prior to the task. Instead, they were given only a task description and asked to generate an appropriate response. This approach evaluates the models' ability to generalize knowledge from their pre-training to this specific context. This strategy was applied to all LLMs. 
    
    \begin{figure*}[ht]
    \centering
    \includegraphics[width=0.8\linewidth]{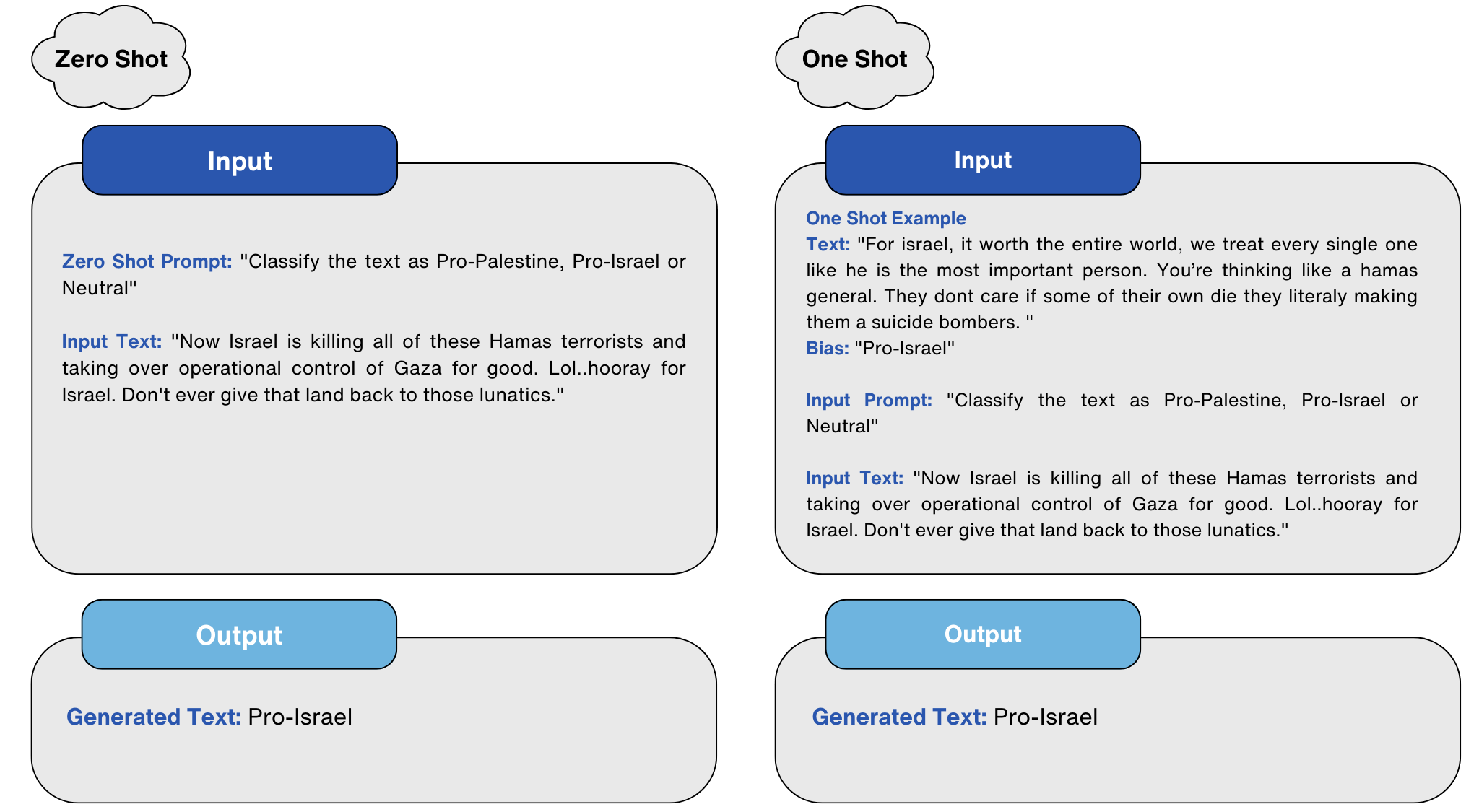}
    \caption{Zero and One Shot Prompting Demonstration}
    \label{fig:zero}
    \end{figure*} 
    
    \item \textbf{One-Shot Prompting:} One-shot prompting involves providing the model with a single example to guide its understanding of the task. This approach leverages the model's capacity to generalize from minimal guidance to improve task performance. This strategy was applied to all LLMs. Figure \ref{fig:zero} illustrates the application of zero and one shot prompting in our study.

    \item \textbf{Few-Shot Prompting:} Few-shot prompting involves providing the model with a small number of examples to help the model learn the task and generalize to new inputs. By offering multiple examples, this strategy allows the model to better understand the task context and the underlying relationships. In our experiments, we specifically used three-shot and five-shot prompting strategies to evaluate the model's performance with additional examples. These strategies were applied to all models except Mixtral, as the API limit on the Huggingface Hub restricted its use. Figure \ref{fig:three} illustrate the application of three-shot and five-shot prompting in our study.

    \begin{figure*}[ht]
    \centering
    \includegraphics[width=0.8\linewidth]{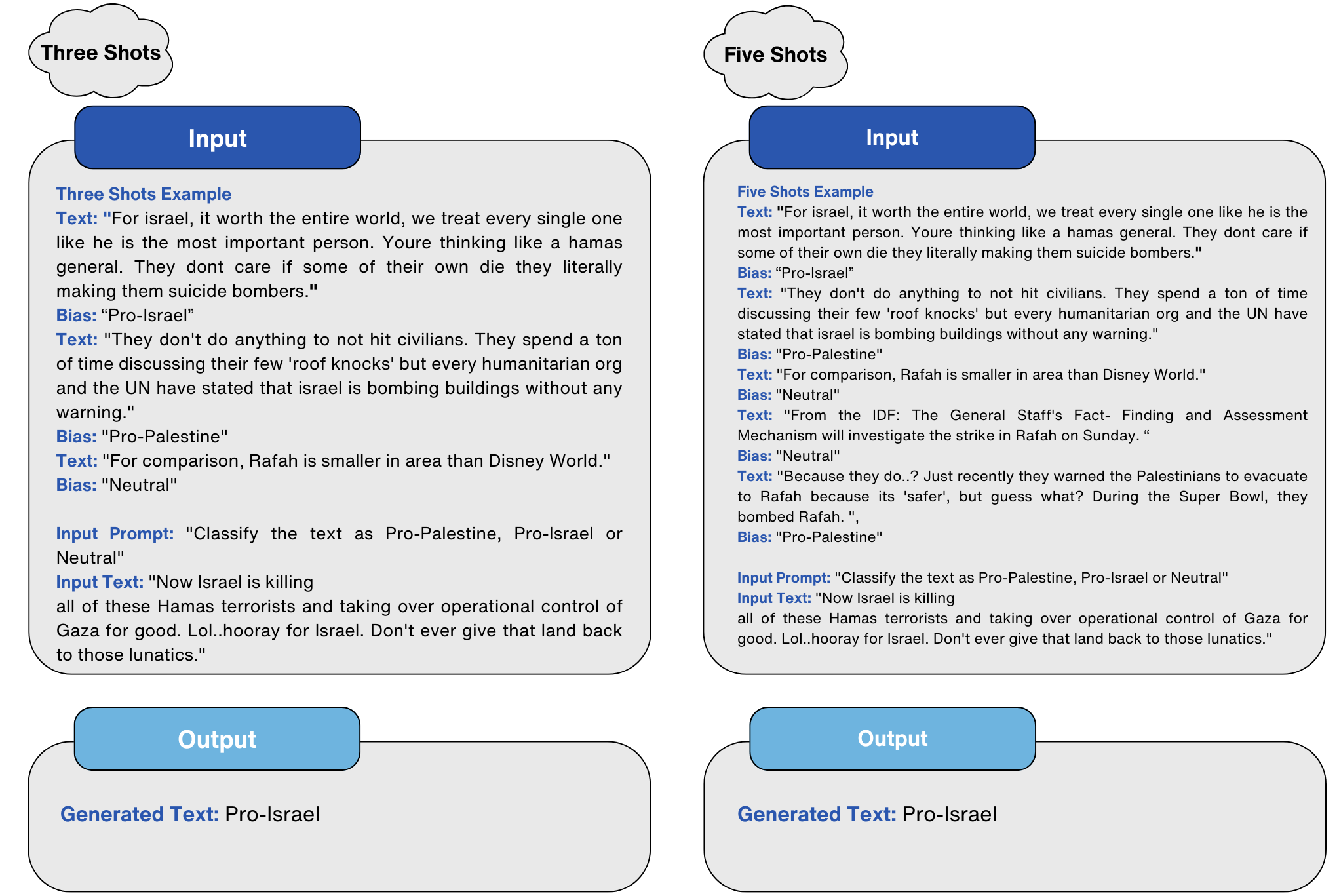}
    \caption{Three and Five Shots Prompting Demonstration}
    \label{fig:three}
    \end{figure*}
    
\end{itemize}

\subsubsection*{\textbf{Novel Prompting and Classification Strategies}}
We used five novel prompting strategies, consisting of Re-read Zero-Shot and One-Shot, Meta-Prompting, Scoring \& Reflective Re-read, and Context Extraction. These methods have not been previously used in this domain. We evaluated these strategies only with Mixtral since its zero- and one-shot performance yielded the best results.

\begin{itemize}
    \item \textbf{Re2 or Re-read:} Re2 or re-read prompting strategy is designed to make the model check its classification and the text itself twice to reassess its initial output. This approach involves careful analysis and minimizes the likelihood of overlooking subtle contextual cues in its initial classification. We extended this approach to a one-shot setting where it reassesses its classification after being presented with an example during the initial classification. This addition combines the iterative strength of Re-read with the guidance provided in a one-shot setting. Figures \ref{fig:re2} and \ref{fig:re2-one} demonstrate the application of the Re-read prompting strategy in our context.
    \begin{figure*}[ht]
    \centering
    \includegraphics[width=0.8\linewidth]{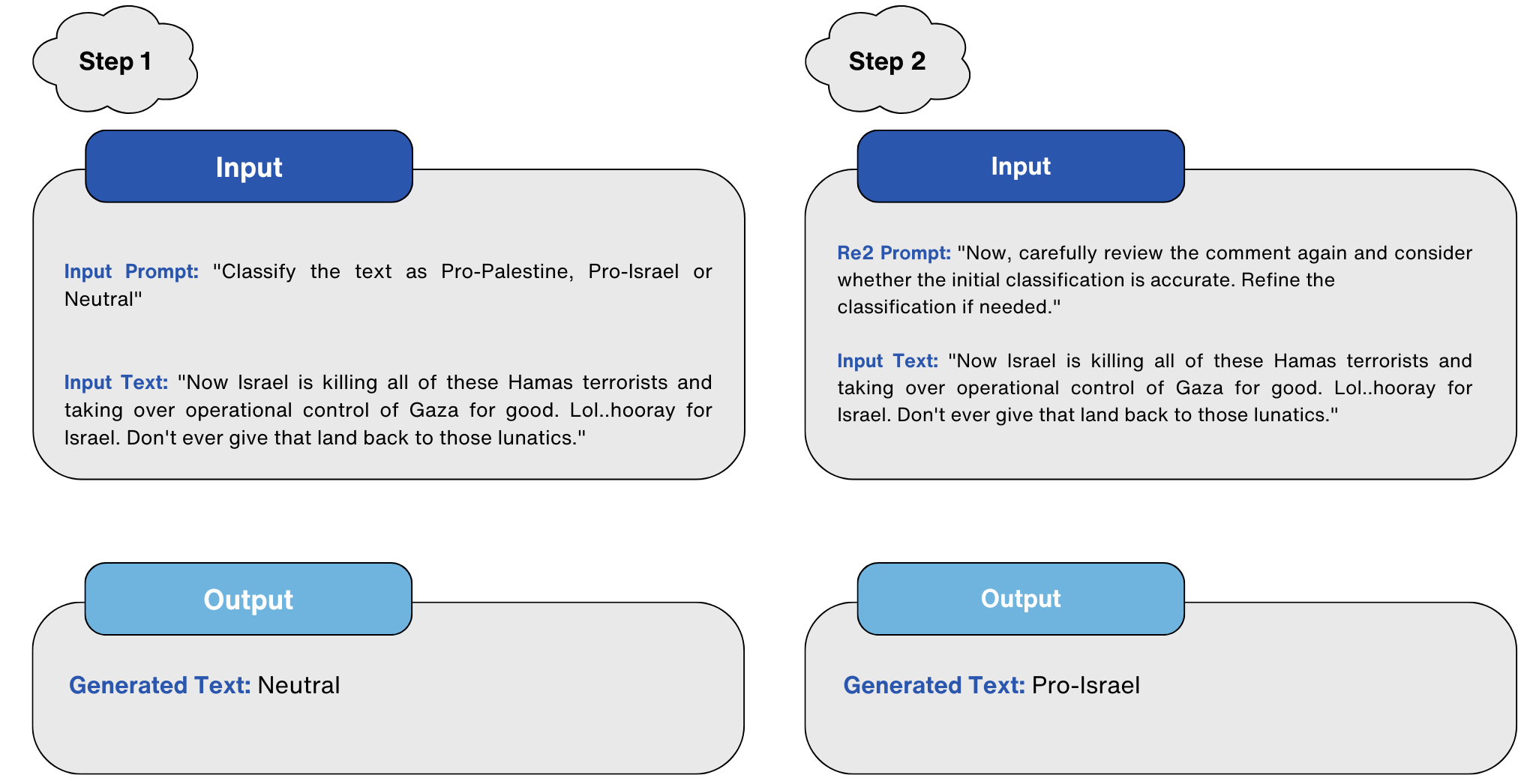}
    \caption{Re-read Prompting Demonstration}
    \label{fig:re2}
    \end{figure*}
    \begin{figure*}[ht]
    \centering
    \includegraphics[width=0.8\linewidth]{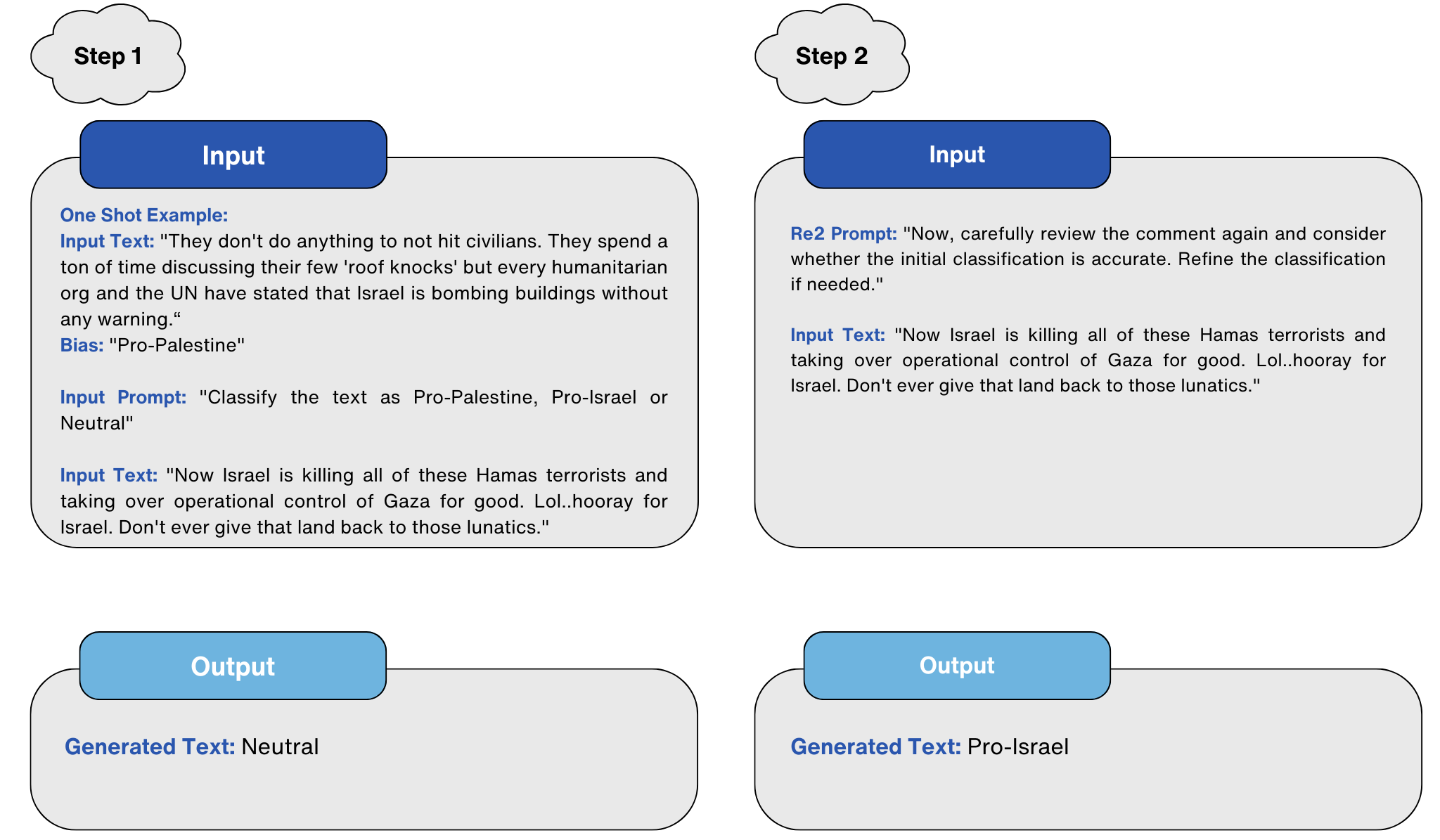}
    \caption{Re-read-One Shot Prompting Demonstration}
    \label{fig:re2-one}
    \end{figure*}  
    \item \textbf{Meta-Prompting (Self-Critique):} Meta-Prompting, in the form of self-critique, refers to a two-step prompting strategy aimed at enhancing the model performance through a refinement process. The approach begins with an initial classification of the input text where it predicts a stance category. After that, the model is asked to critique its own classification and provide a refined final classification. The refined classification can be either the initial classification or a changed one. Figure \ref{fig:meta} demonstrates the use of Meta-Prompting (self-critique) in our context.
    \begin{figure*}[ht]
    \centering
    \includegraphics[width=0.8\linewidth]{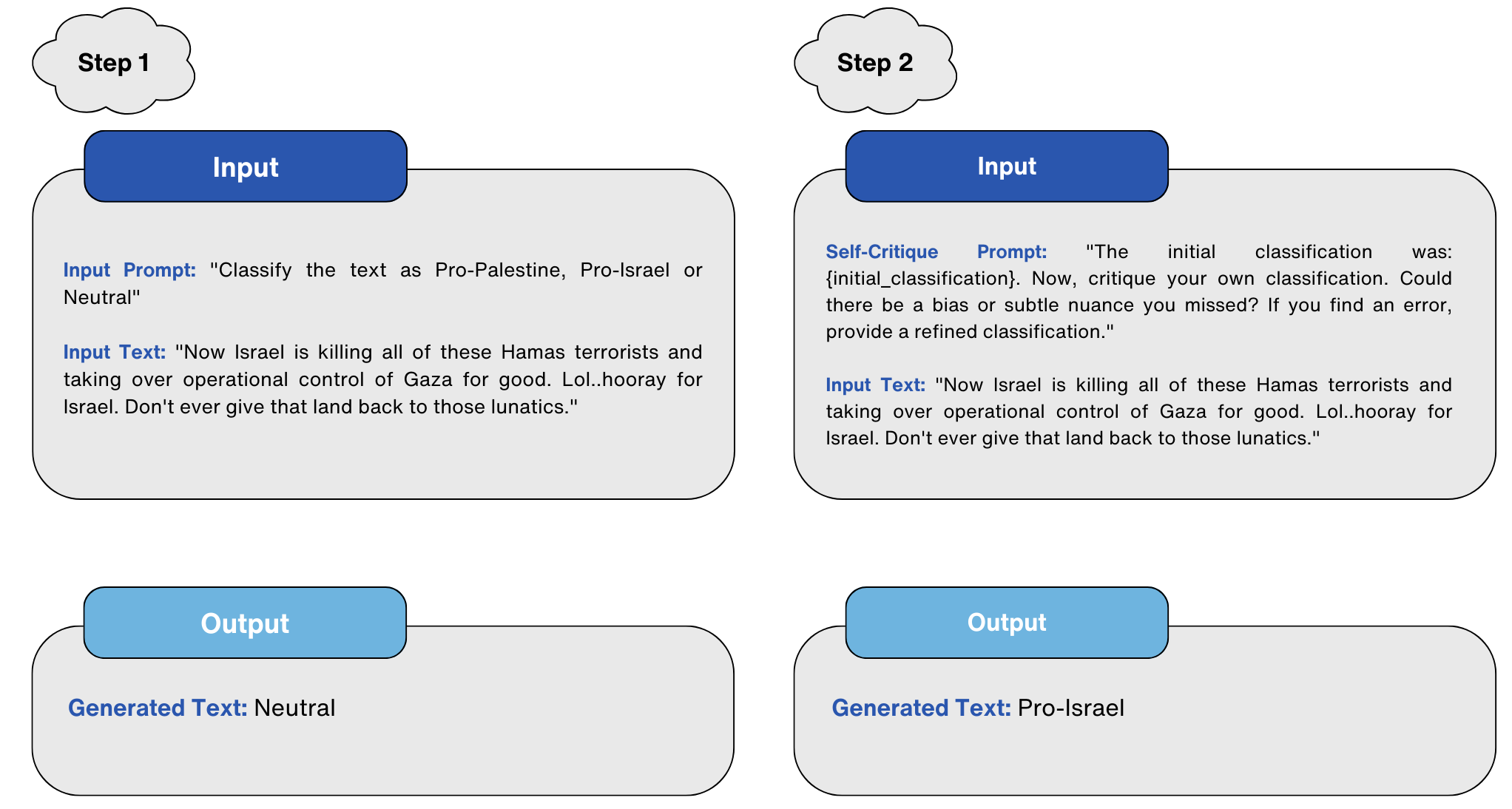}
    \caption{Meta-Prompting (Self-Critique) Demonstration}
    \label{fig:meta}
    \end{figure*} 
    \item \textbf{Context Extraction:} The context extraction strategy was used to identify any contextual elements and embedded stances in the text. The prompt was formulated to be explicit yet open-ended, which may allow the model to effectively capture nuances without adhering to strict guidelines. The prompt allowed for the identification of the textual content first which might allow a better understanding of the potential stance present in it. By keeping the prompt simple and focused, we ensured that the model remained aligned with the primary task of extracting meaningful insights. Figure \ref{fig:context} demonstrates the use of Context Extraction in our context.
    \begin{figure*}[ht]
    \centering
    \includegraphics[width=0.8\linewidth]{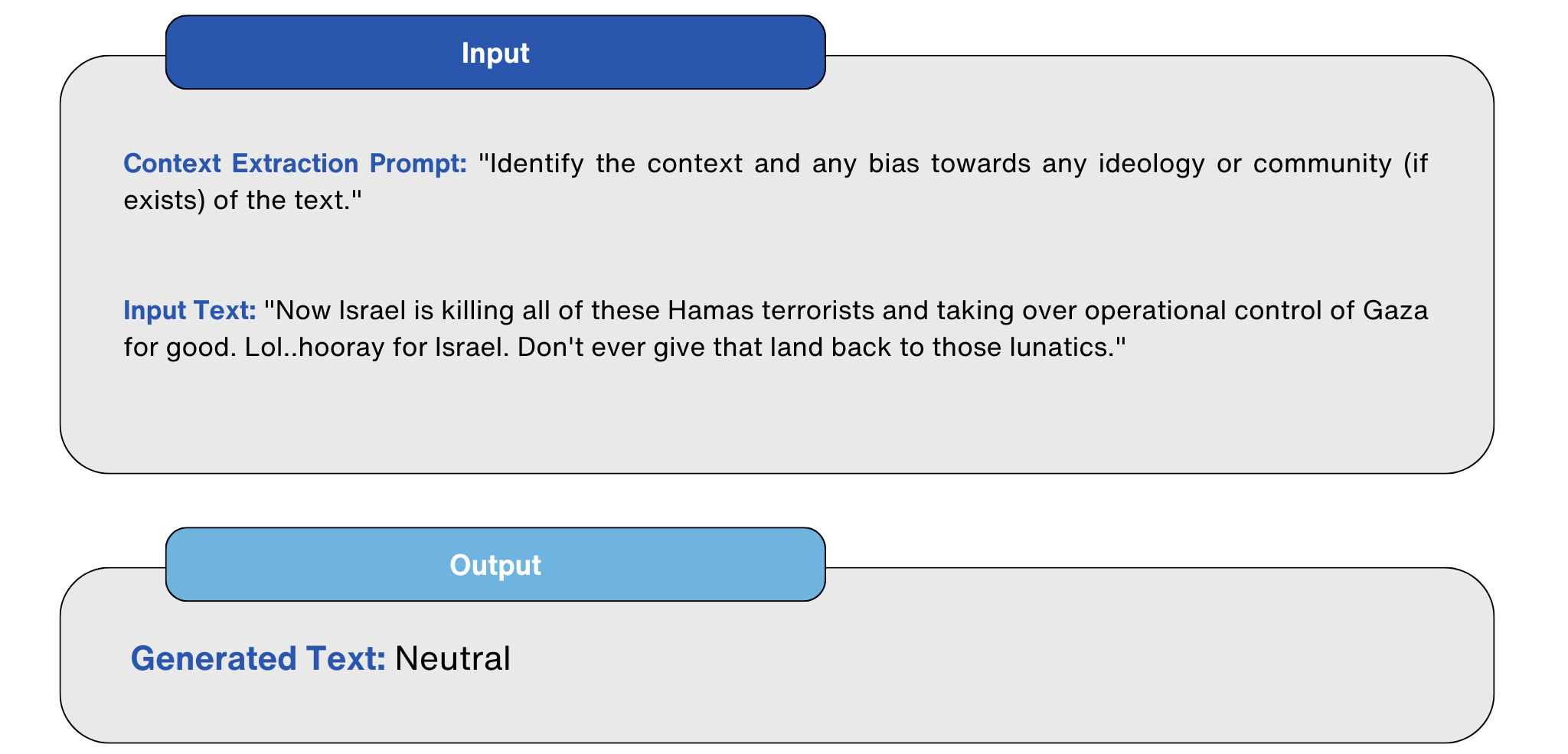}
    \caption{Context Extraction Demonstration}
    \label{fig:context}
    \end{figure*}

    \item \textbf{Scoring and Reflective Re-read:} The Scoring and reflective re-read strategy is a two-step prompting approach designed to encourage the model to assess the input text both qualitatively and quantitatively. The method begins with an initial scoring phase followed by a reflective re-read phase to refine the classification.
    In the initial scoring phase, the model was asked to assign numerical scores (ranging from 1 to 5) to three stance categories: Pro-Israel, Pro-Palestine, and Neutral. This step allows the model to understand the strength of alignment of the text with each category.
    Following the initial scoring phase, the model reevaluates the text and provides a classification based on the initial scores assigned. The scores serve as the primary metric, and the text allows for a further understanding of the context. This allows the model to safeguard against any contextual stances introduced in the first phase while also allowing the model to understand the relative strength of each category for the input text. Figure \ref{fig:score} demonstrates the use of Scoring and Reflective Re-read in our context.
    \begin{figure*}[ht]
    \centering
    \includegraphics[width=0.8\linewidth]{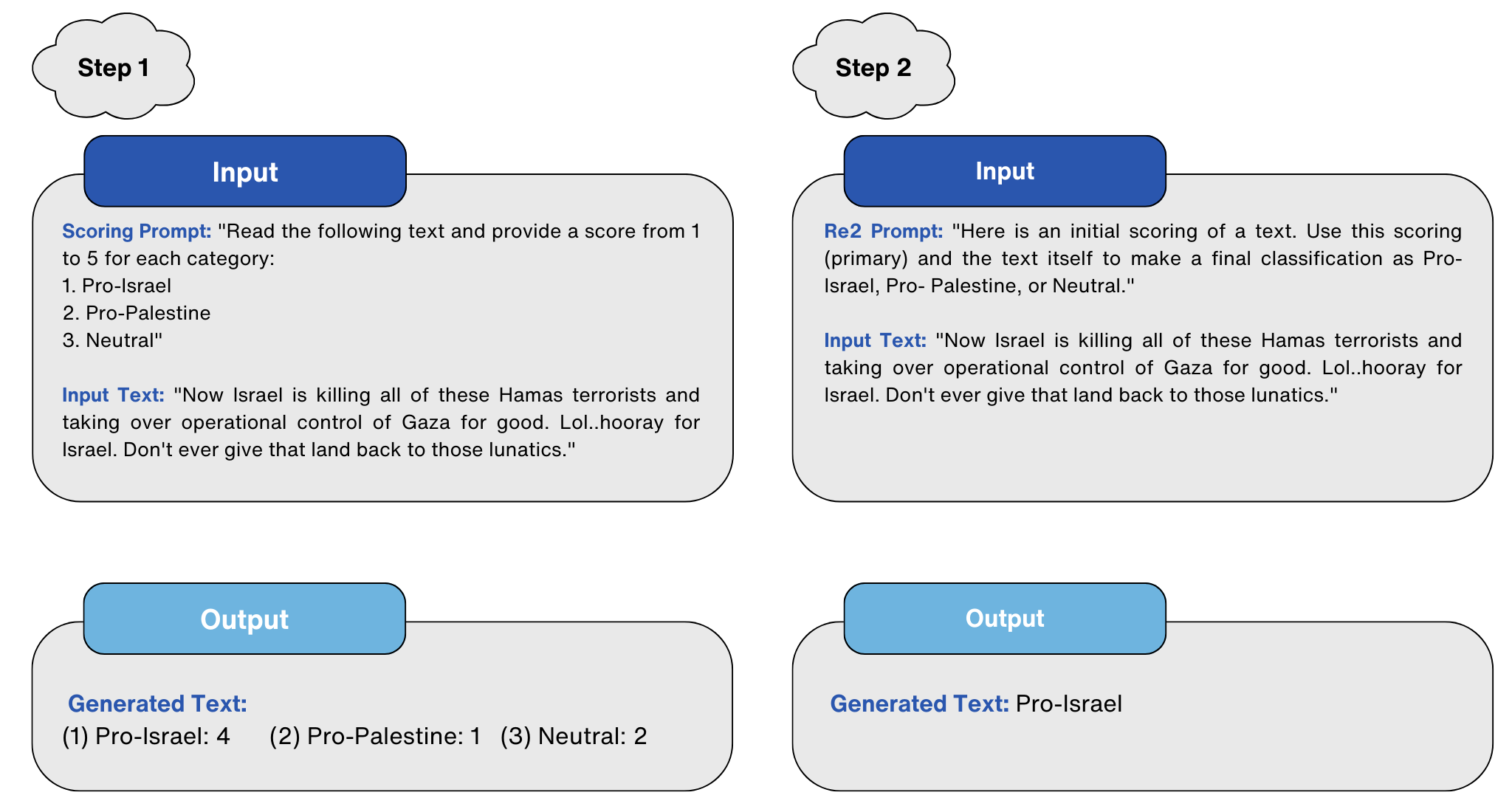}
    \caption{Scoring and Reflective Re-read Demonstration}
    \label{fig:score}
    \end{figure*}
\end{itemize}

\section{Experimental Setup} \label{es}
\subsection{\textbf{Hardware Details}}
The experiments were conducted on the Kaggle and Google Colab platform utilizing the free NVIDIA Tesla P100 and A100 GPU. This facilitated the training and evaluation of the neural network and pre-trained transformer models, particularly in handling computationally expensive processing. Most LLMs, due to their massive sizes, were accessed via the Hugging Face Inference API and OpenRouter API. However, lighter LLMs including Qwen 3-4B, and Falcon 3-3B were downloaded and run locally. 
\subsection{\textbf{Evaluation Metrics}}
To evaluate the performance of the models implemented for sentiment classification, we employed four widely recognized metrics: Accuracy \citep{agustian2024new}, Precision (Macro), Recall (Macro) \citep{cahya2024improving,shakith2024enhancing}, and F1 Score (Macro) \citep{berger2020threshold}. Macro metrics were chosen to ensure a comprehensive analysis, particularly due to the uneven class distribution in the dataset \citep{hote2023open}.  On the other hand, accuracy provides a baseline understanding of model performance across the three stance categories. Detailed descriptions of evaluations metrics are mentioned in Appendix \ref{appendix}.

\begin{table*}[tbp]
\centering
\caption{Performance of models across different architectures. The bold-marked numbers indicate the highest values for each metric.}
\label{tab:model_performance}

\resizebox{\textwidth}{!}{%
\begin{tabular}{@{}llcccc@{}}
\toprule
\textbf{Category} & \textbf{Model} 
& \textbf{Accuracy} 
& \textbf{Precision} 
& \textbf{Recall} 
& \textbf{F1 Score} \\[-0.3ex]
& 
& 
& \textbf{(Macro)} 
& \textbf{(Macro)} 
& \textbf{(Macro)} \\
\midrule 

\multirow{7}{*}{\shortstack{\textbf{Pre-trained} \\ \textbf{Language Models}}} 
    & BERT Cased          & 0.4345  & 0.4442  & 0.4476  & 0.4368  \\
    & BERT Uncased        & 0.5020  & 0.5017  & 0.4358  & 0.4264  \\
    & XLM-RoBERTa         & 0.5314  & 0.5521  & 0.5039  & 0.4452  \\
    & DistilBERT-Uncased  & 0.5695  & 0.5748  & 0.5842  & 0.5649  \\
    & DistilBERT-Cased    & 0.5448  & 0.5315  & 0.5483  & 0.5217  \\
    & ELECTRA-Small       & 0.4251  & 0.5078  & 0.4539  & 0.4236  \\
    & ELECTRA-Base        & 0.4051  & 0.4829  & 0.4715  & 0.4198  \\ 
\midrule 

\multirow{4}{*}{\shortstack{\textbf{Neural} \\ \textbf{Networks}}} 
    & RNN     & 0.4993 & 0.1664 & 0.3333 & 0.2220 \\
    & LSTM    & 0.5354 & 0.3982 & 0.3949 & 0.3387 \\
    & GRU     & 0.5394 & 0.3753 & 0.4159 & 0.3678 \\
    & BiLSTM  & 0.5414 & 0.4053 & 0.4041 & 0.3516 \\ 
\midrule

\multirow{14}{*}{\shortstack{\textbf{Large} \\ \textbf{Language} \\ \textbf{Models}}} 
    & Qwen3-4B Zero Shot     & 0.6583 & 0.6831 & 0.6110 & 0.6250 \\
    & Qwen3-4B One Shot      & 0.6033 & 0.6098 & 0.6054 & 0.6033 \\
    & Qwen3-4B Three Shot    & 0.5600 & 0.5614 & 0.5334 & 0.5391 \\
    & Qwen3-4B Five Shot     & 0.5500 & 0.5618 & 0.5206 & 0.5186 \\
    & Mixtral 8x7B Zero Shot  & 0.6830 & 0.6612 & \textbf{0.6650} & \textbf{0.6628} \\
    & Mixtral 8x7B One Shot   & 0.6802 & \textbf{0.7169} & 0.6555 & 0.6482 \\
    & Mistral 7B Zero Shot    & 0.5207 & 0.6516 & 0.5488 & 0.5212 \\
    & Mistral 7B One Shot     & 0.5174 & 0.6442 & 0.5791 & 0.5109 \\
    & Mistral 7B Three Shots  & 0.5560 & 0.6779 & 0.5942 & 0.5673 \\
    & Mistral 7B Five Shots   & 0.5256 & 0.6625 & 0.5956 & 0.5526 \\
    & Gemma 7B Zero Shot      & 0.4739 & 0.4302 & 0.4344 & 0.4282 \\
    & Gemma 7B One Shot       & 0.6161 & 0.6161 & 0.6161 & 0.6161 \\
    & Gemma 7B Three Shots    & 0.5946 & 0.5432 & 0.5345 & 0.5173 \\
    & Gemma 7B Five Shots     & 0.6219 & 0.5814 & 0.5941 & 0.5901 \\
    & Gemma3-4B Zero Shot    & 0.6583 & 0.6831 & 0.6110 & 0.6250 \\
    & Gemma3-4B One Shot     & 0.6017 & 0.6262 & 0.5794 & 0.5774 \\
    & Gemma3-4B Three Shot   & 0.6733 & 0.7086 & 0.6221 & 0.6230 \\
    & Gemma3-4B Five Shot    & \textbf{0.6867} & 0.7085 & 0.6359 & 0.6437 \\
    & Falcon 7B Zero Shot     & 0.5319 & 0.5569 & 0.5525 & 0.5276 \\
    & Falcon 7B One Shot      & 0.4912 & 0.5549 & 0.5524 & 0.4912 \\
    & Falcon 7B Three Shots   & 0.5434 & 0.5511 & 0.5428 & 0.5288 \\
    & Falcon 7B Five Shots    & 0.5029 & 0.5979 & 0.5029 & 0.5229 \\
    & Falcon3-3B Zero Shot & 0.4900 & 0.4914 & 0.4866 & 0.4886 \\
    & Falcon3-3B One Shot  & 0.5183 & 0.5298 & 0.5476 & 0.5199 \\
    & Falcon3-3B Three Shot & 0.4700 & 0.5168 & 0.4646 & 0.4545 \\
    & Falcon3-3B Five Shot  & 0.4850 & 0.5245 & 0.5030 & 0.4839 \\
\bottomrule

\end{tabular}%
}

\end{table*}

\section{Performance Evaluation} \label{pe}
\subsection{Baseline Models}

The classification performances of all models are listed in Table~\ref{tab:model_performance}. Neural networks, pre-trained language models, and some Large Language Models (LLMs) struggled to achieve an accuracy of 50\%.  Among the pre-trained language models, DistilBERT-Uncased achieved the highest accuracy of 56.95\% with respective macro precision, recall, and F1 scores of 57\%, 58\%, and 56\%. BiLSTM demonstrated the highest accuracy of 54\% among the neural network models while GRU had a greater macro-recall.

Among the LLMs, Mixtral 8x7B, Qwen3-4B, Gemma 7B, and Gemma 3-4B crossed the 60\% accuracy threshold. Gemma 3-4B Five Shot had the highest accuracy, while Mixtral 8x7B Zero and One Shot had the highest macro recall, precision and F1-score. Considering the class imbalance in the dataset, and a comparatively smaller accuracy gain of Gemma 3-4B over Mixtral 8x7B, Mixtral 8x7B is selected as the best performing model for its superiority across macro metrics.

\begin{figure}
    \centering
    \includegraphics[width=0.5\linewidth]{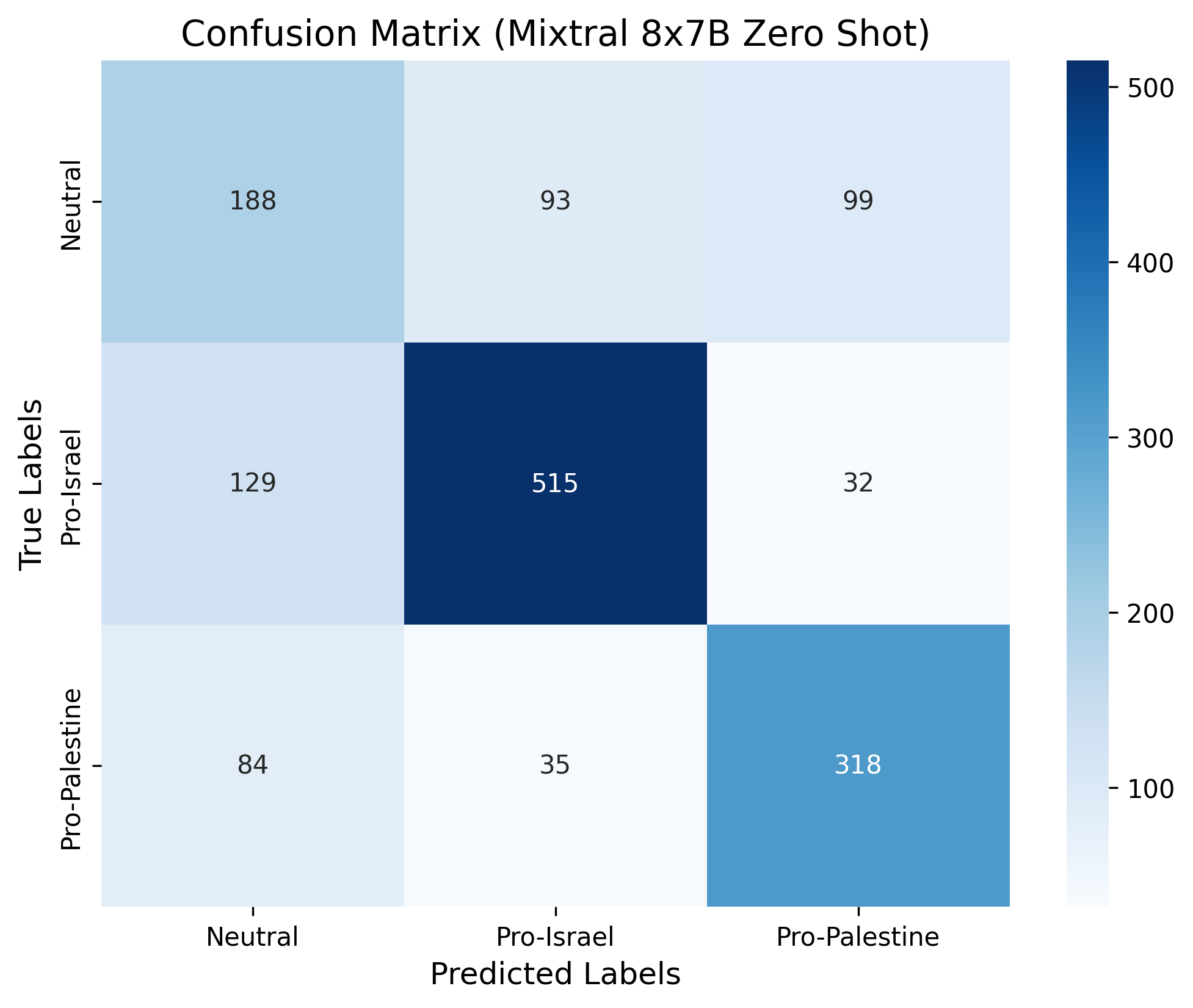}
    \caption{Confusion Matrix of Mixtral 8x7B Zero Shot}
    \label{fig:mixtralzeroshot}
\end{figure}

Table~\ref{tab:class_performance} lists the precision, recall, and F1 score of the best-performing Mixtral 7B Zero Shot. Figure \ref{fig:mixtralzeroshot} shows the confusion matrix. The model was observed to make a significant number of errors in classifying Neutral texts. It showed the best performance in classifying Pro-Israel texts, although the Pro-Palestine classifications were also substantially better than Neutral classifications.
\begin{table}[h]
\centering
\caption{Class-wise performance of Mixtral 8x7B Zero Shot.}
\label{tab:class_performance}
\begin{tabular}{@{}lccc@{}}
\toprule
\textbf{Class} & \textbf{Precision} & \textbf{Recall} & \textbf{F1-Score} \\ \midrule
Neutral & 0.4934 & 0.5068 & 0.5000 \\ 
Pro-Israel & 0.7975 & 0.7621 & 0.7794 \\ 
Pro-Palestine & 0.6928 & 0.7260 & 0.7090 \\ \bottomrule
\end{tabular}
\end{table}

Due to the superiority of the best performing model, we tested the novel prompt engineering techniques on the model to improve its base performance in the next section. We also show the result of LLM majority voting classification in the section after that.

\subsection{Novel Methods}
Table~\ref{tab:ablation_performance} presents the results of the proposed prompting strategies along with the LLM majority voting ensemble for comparison. Among them, Scoring \& Reflective Re-read achieved the highest performance in accuracy with a score of 74.08\%, in macro recall with a score of 73.09\%, and in macro F1 with a score of 72.93\%. However, LLM majority voting had a slightly higher macro precision with a score of 73.52\%. Although the one-shot re-read had an accuracy close to 70\%, it fell short of the Scoring \& Reflective Re-read across all metrics. Considering superiority across all metrics and the difference in performance among the methods, Scoring \& Reflective Re-read emerges as the best performing method.

\begin{table*}[tbp]
\centering
\caption{Performance of Proposed Prompting Strategies and the LLM Majority Voting Ensemble}
\label{tab:ablation_performance}

\resizebox{\textwidth}{!}{%
\begin{tabular}{@{}lcccc@{}}
\toprule
\textbf{Model} & \textbf{Accuracy} & \textbf{Precision (Macro)} & \textbf{Recall (Macro)} & \textbf{F1-Score (Macro)} \\ 
\midrule
Mixtral 8x7B Re2                          & 0.6374 & 0.7246 & 0.6634 & 0.6432 \\ 
Mixtral 8x7B Re2 (One-Shot)               & 0.6965 & 0.7144 & 0.6880 & 0.6896 \\ 
Mixtral 8x7B Meta-Prompting (Self-Critique) & 0.6357 & 0.6086 & 0.6138 & 0.6101 \\ 
Mixtral 8x7B Scoring + Re2                & \textbf{0.7408} & 0.7325 & \textbf{0.7309} & \textbf{0.7293} \\ 
Mixtral 8x7B Context Extraction           & 0.6094 & 0.5969 & 0.6000 & 0.5971 \\ 
LLM Majority Voting & 0.6267 & \textbf{0.7352} & 0.5886 & 0.6311 \\
\bottomrule
\end{tabular}%
}
\end{table*}

\begin{figure}
    \centering
    \includegraphics[width=0.5\linewidth]{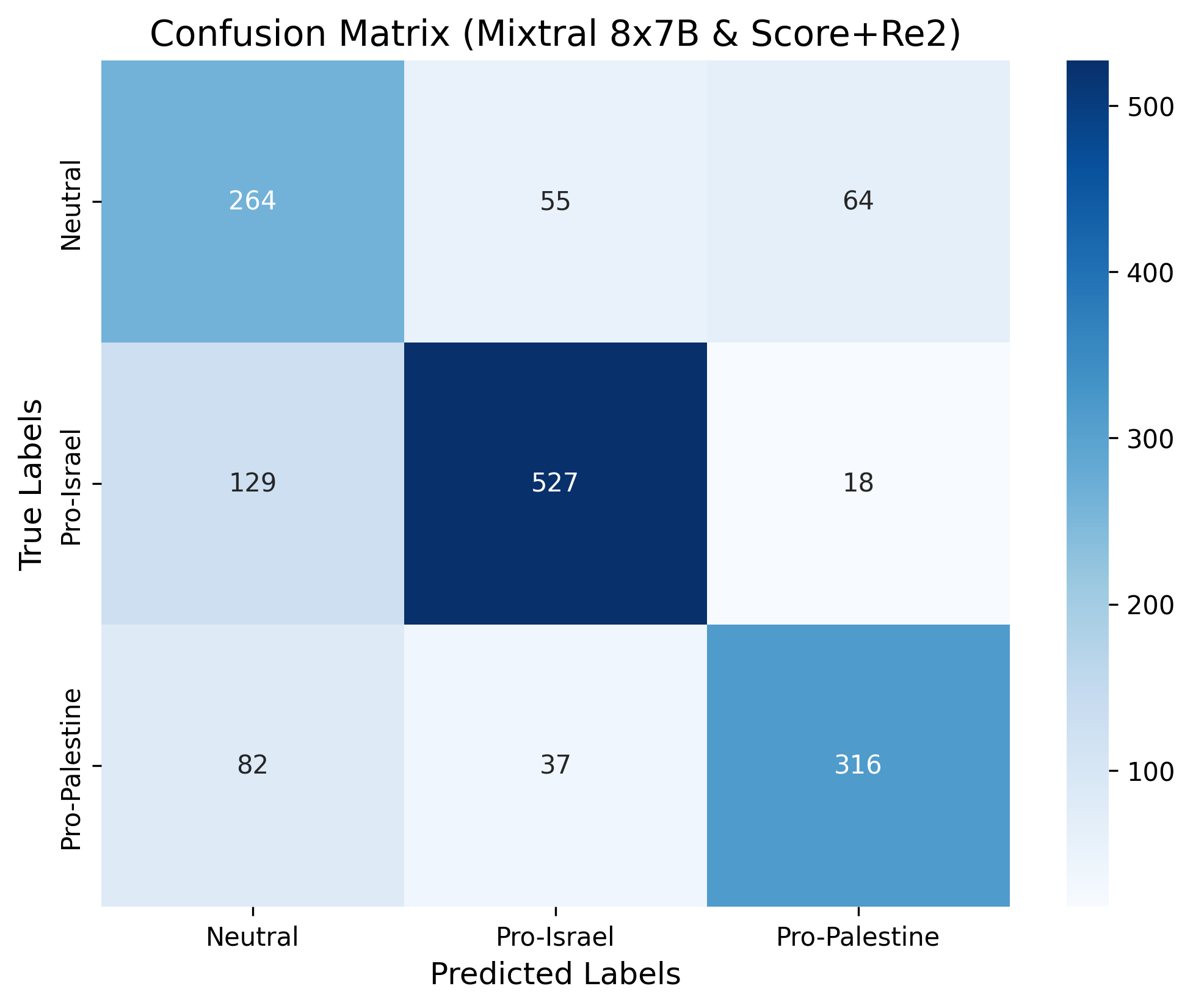}
    \caption{Confusion Matrix (Scoring+Re2)}
    \label{fig:scoringconfusion}
\end{figure}

Table~\ref{tab:class_performance_proposed} lists the class-wise performance of the Scoring \& Reflective Re-read method and Figure \ref{fig:scoringconfusion} shows the confusion matrix. We can observe that the classification performance has increased significantly for all classes compared to the base zero-shot model. The most significant improvements were observed in classifying the neutral classes, where the most error occurred previously. These improvements suggest that the Scoring and Reflective Re-read achieves the best possible outcome out of all tested methods, thus making it our proposed method based on those experimental results. 

\begin{table}[htbp]
\centering
\caption{Class-wise performance of the proposed method.}
\label{tab:class_performance_proposed}
\begin{tabular}{@{}lccc@{}}
\toprule
\textbf{Class} & \textbf{Precision} & \textbf{Recall} & \textbf{F1-Score} \\ \midrule
Neutral & 0.5637 & 0.6769 & 0.6151 \\ 
Pro-Israel & 0.8460 & 0.7847 & 0.8142 \\ 
Pro-Palestine & 0.7879 & 0.7312 & 0.7585 \\ \bottomrule
\end{tabular}
\end{table}

\subsection{LLM Ensemble via Majority Voting}
To examine whether combining predictions from multiple large language models improves classification robustness, we implemented an ensemble strategy based on majority voting. Specifically, predictions from three lightweight LLMs (Gemma 3-4B, Falcon 3-3B, and Qwen 3-4B) were aggregated for each test instance. These models were selected because they can be executed locally and represent diverse architectures among the evaluated LLMs.

For each comment, the stance predictions generated by the three models were collected, and the final label was determined based on the majority class among the predictions. In cases where no majority occurred, the prediction of the best-performing individual model was selected.

The performance of the ensemble model is reported in Table~\ref{tab:ablation_performance}. The majority voting approach achieved an accuracy of 62.67\%, a macro precision of 73.52\%, a macro recall of 58.86\%, and a macro F1-score of 63.11\%. While this ensemble strategy slightly improved macro precision compared to several individual LLM configurations, it did not outperform the proposed \textit{Scoring and Reflective Re-read} prompting strategy across the overall evaluation metrics. The improvement in macro precision indicates that ensemble agreement reduces incorrect positive predictions, leading to more reliable stance assignments.

\subsection{Generalizability Across Subreddits}
To assess generalizability we applied our best-performing method (Mixtral 8x7B + Scoring \& Reflective Re-read) to 1,000 unseen comments from three subreddits (567 from r/IsraelPalestine, 233 from r/worldnews, and 200 from r/Palestine) of the original Kaggle dataset. Table \ref{tab:generalizability} compares the predicted stance distributions with the annotated ground truth from the training set.

\begin{table}[h]
\centering
\caption{Predicted Stance Distributions on Unseen Data Across Subreddits}
\label{tab:generalizability}
\begin{tabular}{@{}l l c c@{}}
\toprule
\textbf{Subreddit} & \textbf{Stance} & \textbf{Annotated (Training)} & \textbf{Predicted (Unseen)} \\
\midrule
\multirow{4}{*}{r/IsraelPalestine} 
& Pro-Israel & 49.2\% (1,565) & 46.0\% (261) \\
& Pro-Palestine & 30.0\% (956) & 38.9\% (220) \\
& Neutral & 20.8\% (662) & 15.1\% (86) \\
& \textbf{Total} & \textbf{3,183} & \textbf{567} \\
\midrule
\multirow{4}{*}{r/worldnews} 
& Pro-Israel & 50.3\% (1,085) & 46.0\% (107) \\
& Pro-Palestine & 15.6\% (337) & 23.2\% (54) \\
& Neutral & 34.1\% (735) & 30.8\% (72) \\
& \textbf{Total} & \textbf{2,157} & \textbf{233} \\
\midrule
\multirow{4}{*}{r/Palestine} 
& Pro-Israel & 6.6\% (38) & 4.0\% (8) \\
& Pro-Palestine & 76.0\% (438) & 91.5\% (183) \\
& Neutral & 17.4\% (100) & 4.5\% (9) \\
& \textbf{Total} & \textbf{576} & \textbf{200} \\
\bottomrule
\end{tabular}
\end{table}

The predicted distributions align with the annotated ground truth, with Pro-Israel comments most frequent in r/IsraelPalestine and r/worldnews, and Pro-Palestine comments dominant in r/Palestine. This offers preliminary evidence that the model might be generalizable, though further validation requires more manually annotated data. 

\section{Broader Impact and Applications} \label{bi}

The implications of this research extend beyond model performance and offer valuable insights into the societal dynamics of polarization in conflict zones. Automated detection of ideological stances can support journalists, policymakers, and humanitarian organizations in monitoring public discourse, identifying emerging narratives, and mitigating the spread of misinformation. In the context of the Israel-Palestine conflict, such tools may facilitate balanced reporting and provide early signals of shifts in public sentiment, which can inform dialogue and peace-building efforts. However, it is to be noted that the proposed prompt engineering strategies may not always generalize reliably across different LLMs or datasets. These tools are also not substitutes for qualitative analysis and hence care should be taken during practical application.

At the same time, the deployment of stance detection models raises critical ethical considerations. Misclassification of sensitive content can unintentionally reinforce existing biases or suppress marginalized voices. Additionally, analyzing social media data during ongoing conflicts requires careful attention to privacy and the potential impact on individuals and communities. To ensure responsible use, we recommend integrating human-in-the-loop validation and interdisciplinary oversight when applying these techniques in politically charged contexts.

\section{Conclusion} \label{conclusion}

This study presents a comprehensive analysis of ideological stance detection in social media discourse surrounding the Israel–Palestine conflict. Using a manually annotated dataset of Reddit comments, we conducted a comparative evaluation of neural networks, pre-trained language models, and large language models (LLMs) across multiple architectures and prompting configurations. In addition, we explored an LLM ensemble approach based on majority voting to assess whether combining predictions from multiple models improves stance classification performance. The findings highlight the nuanced nature of ideological stance expression in conflict-related discussions and underscore the importance of advanced models and prompt engineering techniques for improving classification performance.

Among the evaluated approaches, Mixtral 8x7B combined with the Scoring and Reflective Re-read prompting strategy demonstrated the best overall performance, achieving improvements in accuracy, recall, and F1-score across stance categories. These results illustrate the limitations of traditional models in capturing subtle and implicit ideological cues in politically sensitive discourse and demonstrate the effectiveness of carefully designed prompting strategies. Our analysis also reveals the complexity of polarization in discussions surrounding the Israel–Palestine conflict, where neutral positions proved particularly challenging to detect (50\% precision in baseline models). Furthermore, the stronger performance in detecting Pro-Israel stances compared to Pro-Palestine stances suggests potential asymmetries in how different sides articulate their positions in polarized online environments.

Future work can extend this research by exploring multilingual datasets, incorporating cross-cultural perspectives, investigating fine-tuning approaches for smaller large language models, and examining the ethical implications of ideological stance detection in highly polarized contexts. Overall, the dataset, methodologies, and findings presented in this study contribute to the broader understanding of how natural language processing techniques can be applied to analyze polarization and ideological discourse in conflict-driven social media environments.

\section*{Data Availability Statement}
All data is available on a Github repository: \url{https://github.com/jami78/Conflict-Bias-Eval}.

\section*{Declaration of competing interest}
The authors declare that they have no known competing financial interests or personal relationships that could have appeared to influence the work reported in this paper.

\section*{Acknowledgements}
The authors would like to thank Md Farhan Ishmam for his encouragement and support.

\bibliography{sn-bibliography}

\appendix
\section{Appendix} \label{appendix}
\subsection{Evaluation Metrics}
\textbf{Accuracy:} Accuracy refers to the proportion of correctly predicted instances among the total instances evaluated and serves as an overall indicator of a model's performance 
\begin{equation}
\text{Accuracy} = \frac{TP + TN}{TP + TN + FP + FN}
\end{equation} Where
\begin{itemize}
    \item TP: True Positives
    \item TN: True Negatives
    \item FP: False Positives
    \item FN: False Negatives
\end{itemize}

\textbf{Precision (macro):} Precision evaluates the proportion of correctly predicted instances of a specific class relative to the total predicted instances for that class. Macro-averaging aggregates precision scores across all classes by assigning equal weights to each class, regardless of the number of instances.
\begin{equation}
\text{Macro Precision} = \frac{1}{N} \sum_{i=1}^{N} \frac{TP_i}{TP_i + FP_i}
\end{equation}
Where
\begin{itemize}
    \item N: Total number of classes
    \item \(TP_i\): True positives for the i-th class
    \item \(FP_i\): False positives for the i-th class
\end{itemize}

\textbf{Recall (macro):} Recall, or sensitivity, measures the proportion of correctly identified instances of a class among all actual instances for that class.
\begin{equation}
\text{Macro Recall} = \frac{1}{N} \sum_{i=1}^{N} \frac{TP_i}{TP_i + FN_i}
\end{equation}

Where
\begin{itemize}
    \item \(FN_i\): False negatives for the i-th class
\end{itemize}

\textbf{F1 Score (macro):} The F1 Score combines precision and recall into a single metric by calculating the harmonic mean. Macro F1 averages the F1 scores for all classes. 

\begin{equation}
\text{Macro F1 Score} = \frac{2 \cdot \text{Macro Precision} \cdot \text{Macro Recall}}{\text{Macro Precision} + \text{Macro Recall}}
\end{equation}

Where
\begin{itemize}
    \item \(Precision_i\): Precision for the i-th class
    \item \(Recall_i\): Recall for the i-th class
\end{itemize}

\end{document}